\pdfoutput=1

\documentclass[11pt]{article}

\usepackage[final]{mtsummit25}
\usepackage{copyright}

\usepackage{times}
\usepackage{latexsym}
\usepackage{booktabs}
\usepackage[titletoc]{appendix}

\usepackage[T1]{fontenc}
\usepackage[greek,english]{babel}

\usepackage[utf8]{inputenc}

\usepackage{microtype}

\usepackage{inconsolata}

\usepackage{graphicx}

\title{Gender Bias in English-to-Greek Machine Translation}

\author{Eleni Gkovedarou \\
  Ghent University \\
  \texttt{Eleni.Gkovedarou@ugent.be} \\\And
  Joke Daems \\
  Ghent University \\
  \texttt{Joke.Daems@ugent.be} \\\And
  Luna De Bruyne \\
  University of Antwerp \\
  \texttt{Luna.DeBruyne@uantwerpen.be} \\}

\begin{document}
\maketitle
\begin{abstract}
As the demand for inclusive language increases, concern has grown over the susceptibility of machine translation (MT) systems to reinforce gender stereotypes. This study investigates gender bias in two commercial MT systems, Google Translate and DeepL, focusing on the understudied English-to-Greek language pair. We address three aspects of gender bias: \textit{i)} male bias, \textit{ii)} occupational stereotyping, and \textit{iii}) errors in anti-stereotypical translations. Additionally, we explore the potential of prompted GPT-4o as a bias mitigation tool that provides both gender-explicit and gender-neutral alternatives when necessary. To achieve this, we introduce GendEL, a manually crafted bilingual dataset of 240 gender-ambiguous and unambiguous sentences that feature stereotypical occupational nouns and adjectives. We find persistent gender bias in translations by both MT systems; while they perform well in cases where gender is explicitly defined, with DeepL outperforming both Google Translate and GPT-4o in feminine gender-unambiguous sentences, they are far from producing gender-inclusive or neutral translations when the gender is unspecified. GPT-4o shows promise, generating appropriate gendered and neutral alternatives for most ambiguous cases, though residual biases remain evident. As one of the first comprehensive studies on gender bias in English-to-Greek MT, we provide both our data and code at \url{https://github.com/elenigkove/genderbias_EN-EL_MT}.

\end{abstract}

\section{Introduction}
As the demand for inclusive language grows, the role of technology in shaping linguistic norms becomes increasingly important. While MT is widely used for communication, cost reduction, and accessibility \citep{nurminen_machine_2020, vieira_machine_2020, asscher_human_2023} and despite advancements in the field with state-of-the-art neural machine translation (NMT) systems, they often produce inaccurate, ungrammatical, or biased translations, particularly in assigning gender \citep{stanovsky-etal-2019-evaluating, currey-etal-2022-mt}. Concern has therefore grown over the susceptibility of those systems to translate based on gender stereotypes and the perpetuation of such biases via them, as they can have real-word, harmful consequences for users and society \cite{monti_20, savoldi-etal-2021-gender, lardelli_23}.

A MT model is considered biased ``when it \textit{systematically} and \textit{unfairly} discriminates against certain individuals or groups in favour of others'' \cite{friedman_1996}. 
While human translators rely on the wider context to determine the appropriate gender, most current MT systems do not; instead, they rely on spurious correlations in the (biased) training data which often lead to defaulting to either male or female forms \citep{vanmassenhove-etal-2018-getting, kocmi-etal-2020-gender}. These biases reflect the gender stereotypes that are present in our society. As \citet{saunders20} highlight, translations are more accurate for sentences involving men due to the training data naturally featuring men more than women, as well as for sentences that align with stereotypical gender roles. For example, references of ``male doctors'' are more reliably translated than those of ``male nurses'' \citep{sun-etal-2019-mitigating}, while more errors are detected when the source texts exhibit anti-stereotypical professions, e.g. ``female doctors'' or ``male nurses'' \citep{kocmi-etal-2020-gender}.

This study focuses on gender bias that occurs when translating from English, a notional gender language where gender is not always specified, into Modern Greek (henceforth Greek), a grammatical gender language where it is morphologically and semantically necessary to mark the gender \citep{savoldi-etal-2021-gender, currey-etal-2022-mt}. While prior work has focused on high-resource languages such as German, Spanish, and French \citep{currey-etal-2022-mt, Zhao:24, lee24, Lardelli:24}, Greek remains understudied despite preliminary exploration on document-level \citep{karastergiou24} and labor-domain bias analysis \citep{Mastromichalakis:24}.

These cross-linguistic differences can lead to ambiguities that are difficult to resolve, especially for sentence-level MT systems \citep{vanmassenhove-etal-2021-neutral}, making it more complex to accurately assign gender or maintain gender neutrality. This raises important questions as to what constitutes appropriate handling of gender-ambiguous inputs. Although individual translations (e.g., translating ``the worker'' as either feminine or masculine) may be grammatically valid, aggregate patterns reveal systemic biases. We argue that ideal MT systems should recognise when gender cannot be determined from context and provide either gender-neutral options, or a range of plausible gendered interpretations. This framework informs our evaluation of existing systems and our proposed LLM-based mitigation approach.

The main contribution of this work includes:
\begin{enumerate}
  \item We release \textbf{GendEL}, the first (handcrafted) dataset for evaluating English-to-Greek translations, which comprises \textit{i)} 240 gender-ambiguous and unambiguous English sentences, \textit{ii)} human-generated Greek alternate translations (feminine/masculine/neutral variants), and \textit{iii)} non-binary representations (singular `they').
  \item We focus on the under-represented English-to-Greek language pair and demonstrate that gender bias is persistent in translations by \textbf{Google Translate}\footnote{\url{https://translate.google.com/}} and \textbf{DeepL}.\footnote{\url{https://www.deepl.com/en/translator/}}
  \item We explore the potential of prompting a large language model (LLM), \textbf{GPT-4o}\footnote{\url{https://openai.com/index/hello-gpt-4o/}}  (OpenAI, 2024), to help in bias mitigation by generating not only acurrate gender assignments, but also gender-explicit and gender-neutral alternatives for ambiguous English sentences.
\end{enumerate} 

\paragraph{Bias Statement} In this paper, we analyse English-to-Greek MT outputs to study stereotypical gender associations with professional occupations and adjectives. We analyse the presence of three patterns of gender bias and specifically address the issue of \textbf{representational harm} \citep{blodgett_20}. Representational harm is categorised into two types: \textit{under-representation}, which reduces the visibility of certain social groups (such as women and non-binary individuals), and \textit{stereotyping}, which reinforces negative generalisations (e.g., associating women with less prestigious professions compared to men) \citep{savoldi-etal-2021-gender}.

\section{Related Work} \label{Related_Work}
\subsection{Greek as a grammatical gender language}
Understanding Greek's grammatical gender system is important for analysing gender bias in translation, as the language's structural requirements and gender-inclusive techniques influence the manifestation of gender in the outputs. Greek has three declensions: masculine, feminine, and neuter. Masculine and feminine typically mark human referents, while neuter is mostly assigned to inanimate objects, though certain neuter nouns such as ``\textgreek{το άτομο}'' (the individual) or ``\textgreek{το μέλος}'' (the member) refer to humans. In general, nouns denoting male human beings are grammatically masculine and nouns denoting female human beings are grammatically feminine \citep{pavlidou_grammatical_2004}. \citet{alvanoudi15book} points out that Greek's deeply embedded grammatical gender is restrictive, as gender marking must be encoded in most of the speakers' utterances.

In this study, the focus is primarily on occupational nouns, which often have overt gender marking through suffixes, e.g. ``\textgreek{δάσκαλος}'' (teacher [M]) and ``\textgreek{δασκάλα}'' (teacher [F]). However, gender marking can also be covert, known as \textit{common gender} or \textit{epicene} nouns, which share the same form for male and female referents. In these cases, disambiguation relies on articles or other modifiers; for instance, ``\textgreek{ο/η δικηγόρος}'' (the [M/F] lawyer) uses a clearly masculine suffix for either gender, while the gender is clarified only by the article. The morphological formation and choice of suffix for occupational nouns can be semantically linked to implicit connotations and is an indication of linguistic sexism \citep{sarri24}. Several proposals \citep{triantaf63, tsopanakis82, tsokalidou96, gkasouka2018, sarri24} have been made to feminise such epicenes in ways that align with the morphological and inflectional system of Greek while adhering to grammatical gender agreement, e.g. ``\textgreek{η δικηγορίνα}''. However, despite a slow increase in acceptance and usage, most of these feminised terms have not become standardised in official language use.

In some instances, both masculine and feminine forms exist for occupational nouns, but the feminine versions often carry semantic, stylistic, or register differences. These differences can potentially result in negative connotations or reduced social weight. For example, ``\textgreek{δήμαρχος}'' (mayor) and ``\textgreek{δημαρχέσα}'' (the wife of a mayor or a female mayor) differ in both gender marking and societal implications \citep{kalfadopoulou22}.

\subsection{Gender-inclusive practices in Greek}
\citet{pavlidou_grammatical_2004} found that Greek vocabulary is male-dominated, with masculine nouns for human reference nearly doubling feminine ones. This male bias is evident also in the use of the generic masculine, i.e. using the masculine form of a term even when referring to mixed-gender groups, which erases women and non-binary individuals by treating male experience as ``the default'' \citep{mucchi05}. To address this, Greek institutions have proposed gender-inclusive strategies, such as 1) combined forms (``\textgreek{ο καθηγητής} / \textgreek{η καθηγήτρια}''), 2) combined suffixes (``\textgreek{ο/η καθηγητής/τρια}''), and 3) exclusive feminine forms for female entities. However, these approaches assume a binary gender framework, implying that references to the feminine and masculine are supposedly exhaustive \citep{ntouvlis20}. Unlike English, which has adopted the singular `they', Greek language reform has progressed at a slower pace, and equivalent options are not available, yet.

The most accessible inclusive mechanism in Greek today, also included in the aforementioned guides, is \textbf{gender-neutral language} and can be achieved through techniques like passive syntax, second-person plural, imperatives, circumlocution, and neuter grammatical gender \citep{kalfadopoulou22}. Each technique has its limitations and is context-dependent, making application particularly challenging. For our study on occupational nouns, circumlocution and neuter forms are the most viable techniques. As noted by \citet{piergentili23}, gender-neutral rephrasings and synonyms is a workable paradigm toward more inclusive MT when gender is unknown or simply irrelevant. For example, gendered ``\textgreek{καθηγητές/καθηγήτριες}'' (professors [M/F]) can be replaced with neutral expressions like ``\textgreek{το διδακτικό προσωπικό}'' (the teaching staff) or ``\textgreek{τα μέλη του διδακτικού προσωπικού}'' (the members [N] of the teaching staff). In this way, we restructure sentences to eliminate gendered language, adopting neuter terms like ``\textgreek{το άτομο}'' (the individual) or ``\textgreek{το μέλος}'' (the member). Such neuter forms are also used by non-binary individuals, alongside neologisms like ``\textgreek{το φίλο}'' (the friend), which, as any newly coined word, is rather far from earning mainstream acceptance. Finally, with regard to written discourse, gender-neutral symbols like @ (used as a suffix, e.g. ``\textgreek{τ}@ \textgreek{φίλ}@'') are increasingly becoming popular on social media \citep{ntouvlis20}.

\subsection{Gender bias in MT \& LLMs as gender rewriters}
In MT, we document previous research on gender bias focused on coreference resolution and pronoun translation in relation to human entities \citep{rudinger18, zhao18, prates19, cho19, stanovsky-etal-2019-evaluating, kocmi-etal-2020-gender, gonen20, levy21, currey-etal-2022-mt, robinson24}. The analyses show that popular MT systems are significantly prone to perpetuate but also exacerbate biases through systematic gender-related translation errors, while underlining the challenges of gender bias mitigation. 

Approaches to this problem have involved training models from scratch on artificially gender-balanced datasets \citep{zhao18, zmigrod19}, using debiased embeddings \citep{bolukbasi16, escude19}, and annotating data with speakers' gender information \citep{vanmassenhove-etal-2018-getting}. Additional methods include POS tagging \citep{elaraby18}, word-level gender tagging \citep{stafanovics20, saundersbill20}, fine-tuning \citep{saunders20}, or gender re-inflection of references such as Google Translate \citep{johnson20} and Fairslator{\footnote{\url{https://www.fairslator.com/}} \citep{mechura22}. However, as \citet{savoldi-etal-2021-gender} highlight, there is no definitive, state-of-the-art solution for mitigating bias in MT; instead, these interventions typically address isolated aspects of the problem with targeted, modular solutions.  It is also worth noting that most of these studies largely operate within a binary framework, emphasising masculine and feminine forms into grammatical languages, which ultimately limits their inclusivity.

With the advent of artificial intelligence (AI), the translation capabilities and potential of LLMs in bias mitigation are being explored. \citet{ghosh23} showed that GPT-3 tends to reinforce stereotypes and struggles with gender-neutral pronouns, as it systematically converts them into binary forms between English and Bengali, as well as five other low-resource languages (Farsi, Malay, Tagalog, Thai, Turkish). In a related study, \citet{Vanmassenhove24} observed a strong male bias in GPT-3.5's English-Italian translation, despite being explicitly instructed to provide all possible gender alternatives. \citet{lee24} tested prompting GPT-3.5 Turbo and Llama 2 70b Chat{\footnote{\url{https://huggingface.co/meta-llama/Llama-2-70b-chat-hf}}} for English to Spanish, French, and Italian translations, and despite showing promise for controlled gendered outputs, their approach remained limited to binary representations.

Another study by \citet{sanchez24} tested few-shot prompting on Llama-7B{\footnote{\url{https://huggingface.co/meta-llama/Llama-2-7b}}} and showed a sufficient level of control over binary gender forms in 25 languages, underlining that similar strategies could be applicable to non-binary alternatives. \citet{piergentili24} extended this approach using the Neo-GATE dataset that incorporates non-binary structures and found that GPT-4 and Mixtral{\footnote{\url{https://huggingface.co/docs/transformers/en/model_doc/mixtral}}} performed best with few-shot prompting for English-Italian translations.

LLMs like GPT-4 have shown potential for gender-neutral translations when prompted with few-shot exemplars, though zero-shot performance remains inadequate \citep{savoldi24-prompt}. \citet{rarrick24} developed a translation-rewriting solution with GPT-4, using chain-of-thought prompting, which involved explicitly providing the LLM with step-by-step reasoning and detailed clarifications in the examples. The results indicate that, while the model achieved high accuracy in pronoun adjustments, it struggled with gendered nouns, showing a limitation in handling complex gender adjustments.

\section{Methodology}
\subsection{Dataset Preparation}
We created GendEL, a manually crafted dataset of 240 gender-ambiguous and gender-unambiguous English sentences, based on a list of 40 occupational nouns. For each occupation, we created a subset of six sentences: a baseline (`ambiguous base') and five variations, which modify the baseline in a specific way (e.g. by adding a stereotypical adjective or a pronoun). The baseline template is: \textit{The [OCCUPATION] finished the work}.\footnote{Inspired by \citet{saundersbill20}, who used binary-gendered examples like ``The actor finished her/his work'' for gender debiasing in translation.}

The occupational nouns were selected from \citet{troles21}, using data from the US Bureau of Labor Statistics (2019). Occupations were classified as male- or female-biased if over 50\% of workers were men or women, respectively; e.g. 93\% of carpenters are men, 80\% of librarians are women  (\hyperref[sec:appendixA]{Appendix A}). We created 20 subsets for male- and 20 for female-biased occupations.

To enrich the scope of the study, five additional sentence types were derived from the `ambiguous base', resulting in six types overall (examples are provided in \hyperref[tab:1]{Table 1}):

\begin{itemize}
    \item \textbf{Ambiguous + male-biased adj.}: Ambiguous sentence with male-biased adjective.
    \item \textbf{Ambiguous + female-biased adj.}: Ambiguous sentence with female-biased adjective.
    \item \textbf{Unambiguous [Male]}: Unambiguous sentence with a masculine pronoun.
    \item \textbf{Unambiguous [Female]}: Unambiguous sentence with a feminine pronoun.
    \item \textbf{Ambiguous / unambiguous [Non-binary]}: Uses singular `they', which makes the sentence either gender-ambiguous (gender is purposefully omitted or undefined) or unambiguous (referring to a non-binary individual).
\end{itemize}

\begin{table*}
\centering
\begin{tabular}{@{}ll@{}}
\toprule
\textbf{Sentence Type} & \textbf{Example} \\ \midrule
ambiguous base & The assistant finished the work. \\
ambiguous + male-biased adj. & The \textit{eminent} assistant finished the work. \\
ambiguous + female-biased adj. & The \textit{sassy} assistant finished the work. \\
unambiguous {[}Male{]} & The assistant finished \textit{his} work. \\
unambiguous {[}Female{]} & The assistant finished \textit{her} work. \\
ambiguous / unambiguous {[}Non-binary{]} & The assistant finished \textit{their} work. \\ \bottomrule
\end{tabular}
\caption{Examples of sentence types representing a subset (for occupational noun ``assistant'') from GendEL.}
\label{tab:1}
\end{table*}

The gender-biased adjectives were sourced from \citet{troles21}, selecting 10 male- and 10 female-biased adjectives that we evenly and randomly distributed across the dataset (\hyperref[sec:appendixA]{Appendix A}). These sentences were kept short and simple, minimising linguistic diversity. This ensured that there were no factors influencing the referents' gender other than the investigated words, i.e. occupational nouns, pronouns, and adjectives.}

All sentences were manually translated into Greek by the author of this study. Ambiguous sentences received three translations (masculine, feminine, neutral), while unambiguous ones had a single correct translation. For the `ambiguous~/ unambiguous [Non-binary]', we consider that there is only one correct translation which is a gender-neutral one. A sample of GendEL, including male- and female-biased subsets, is in \hyperref[sec:appendixB]{Appendix B}.

\subsection{Translation Systems}
We first test two widely used commercial MT models: Google Translate and DeepL. Both of these systems have implemented a feature that provides two outputs for short gender-ambiguous queries. However, while Google Translate offers this feature for some languages, Greek is not among the supported languages for gender-ambiguous sentence outputs. On the other hand, DeepL provides this feature for Greek, but its implementation is inconsistent across different sentence structures and contexts. All sentences of GendEL were translated with both MT systems. For gender-ambiguous sentences we reported the main output and any gender alternatives provided by DeepL.

In addition to these models, we included GPT-4o into our analysis to explore its potential for gender-inclusive translations. The advantage of using an LLM is that it can be directed using a prompt, allowing for customised outputs. To achieve this, we leveraged OpenAI's GPT-4o\footnote{Model version: gpt-4o-2024-08-06} via the OpenAI API and, similarly to the approach of \citet{rarrick24}, we used few-shot chain-of-thought prompting to encourage the model to produce gender rewrites when necessary, i.e. in gender-ambiguous cases. The full prompt is available in \hyperref[sec:appendixC]{Appendix D}.

\subsection{Annotation}
For the finalisation of the dataset, we manually annotated the translations generated by the two MT systems and LLM with labels to indicate the gender representation in the output. Particularly, the labels included \textbf{M} (masculine), \textbf{F} (feminine), \textbf{N} (neutral), or combinations thereof, such as \textbf{M-F-N}, \textbf{M-F}, and \textbf{M-N}, to capture cases with alternative translations. Additionally, we introduced four distinct error labels to classify certain issues:

\begin{itemize}
    \item \textbf{error [1]}: Incorrect or nonsensical translation (e.g., non-existing words, singular replaced with plural etc.).
    \item \textbf{error [2]}: Mixed genders in translation (e.g., masculine noun with feminine pronoun: ``The \textit{male} farmer finished \textit{her} work.'').
    \item \textbf{error [3]}: Erroneous or insufficient neutralisation techniques (e.g., using double forms or epicene nouns which imply binary gender and do not reflect true gender neutrality).
    \item \textbf{error [4]}: Adjective missing from the translation.
\end{itemize}

\subsection{Evaluation}

The evaluation of the models' outputs was performed using a mixed-methods approach. Automatic evaluation metrics such as BLEU \citep{papineni02} or TER \citep{snover06} will not be included due to their known limitations. These metrics, while commonly used to evaluate translation accuracy, treat all errors equally and lack sensitivity to certain linguistic phenomena, such as gender bias \citep{sennrich17}. Following \citet{freitag_experts_2021}, we acknowledge that human evaluation remains the gold standard for evaluating translation quality, and we therefore rely on it to assess the translations' validity and alignment with gender-inclusive practices in Greek. 

\subsubsection{Gender Bias in MT Systems}
To investigate the presence of gender bias in the MT systems, we examined three key patterns of bias: \textit{i)} \textbf{male bias}, \textit{ii)} \textbf{occupational stereotyping}, and \textit{iii)} \textbf{errors in anti-stereotypical gender assignments}.

\paragraph{Male Bias} We tested whether Google Translate and DeepL exhibit a tendency to default to masculine forms when translating gender-ambiguous English sentences into Greek. To evaluate this, we calculated the distribution of gendered outputs exclusively for the ambiguous sentences,\footnote{`ambiguous base', `ambiguous + male-biased adj.', `ambiguous + female-biased adj.', `ambiguous + unambiguous [Non-binary]'} where no explicit cues were provided in the source text. By analysing these trends, we aimed to identify systematic male bias in the systems' translation behaviour.

\paragraph{Occupational Stereotyping} We examined whether the MT systems reinforce traditional gender roles associated with specific professions (e.g. ``male doctor'', ``female nurse''), focusing again on the gender-ambiguous sentences. The frequency of stereotyping was calculated for male- and female-biased occupations, with statistical significance tested using Fischer's exact test \citep{fisher_92}.

\paragraph{Anti-Stereotypical Gender Assignments} We analysed the outputs of gender-unambiguous English sentences\footnote{`unambiguous [Male]', `unambiguous [Female]'} comparing error rates between anti-stereotypical (e.g. ``female doctor'') and stereotypical cases (e.g. ``male doctor''). Fischer's exact test was used, followed by qualitative analysis of significant cases.

\subsubsection{GPT-4 on Gender Bias Mitigation}
We evaluated prompted GPT-4o's ability to produce gender-inclusive translations. First, we calculated the gender distribution across all sentence types of GendEL. Second, we analysed the error distribution and conducted a qualitative review to identify factors influencing the model's performance and highlighted areas where it deviated in terms of gender-inclusive practices.

\section{Results}
A preliminary analysis reveals substantial differences in gender and error distributions across the three models (\hyperref[tab:2]{Table 2}). Google Translate and DeepL strongly favoured masculine forms (65.8\% and 63.3\%), with significantly fewer feminine translations (16.2\% and 25.8\%). In contrast, prompted GPT-4o showed a more balanced approach, generating 16.7\% masculine, 15.8\% feminine, and 12.9\% neutral translations. For 42.9\% of sentences it generated three gendered alternatives (M-F-N). These results align more closely with the gold standard, which aims for equal representation of gendered and neutral translations (16.7\% for each gender and 50\% for alternatives). The gold standard highlights the gap between observed and ideal distributions, particularly the lack of gender-neutral forms and alternatives by Google Translate and DeepL.

Regarding errors, Google Translate had the highest rates: 13.3\% of translations included mixed genders, 3.7\% were incorrect/nonsensical translations, and 0.4\% omitted the adjective. DeepL performed better, with only 1.3\% mixed-gender and 1.3\% incorrect/nonsensical translations. GPT-4o also showed low error rates (1.7\% mixed-gender, 1.3\% incorrect/nonsensical) but omitted adjectives in 4.2\% of cases. Interestingly, GPT-4o introduced a unique error type; as the only model that actively attempted to provide neutral forms, it did not always succeed, resulting in 3.3\% of cases that contained errors related to the neutralisation techniques.

\begin{table*}
\centering
\begin{tabular}{@{}cccccc@{}}
\toprule
\textbf{Label} & \textbf{Google Translate} & \textbf{DeepL} & \textbf{Prompted GPT-4o} & \textbf{Gold standards} \\ 
\midrule
\textbf{M} & 158 (65.8\%) & 152 (63.3\%) & 40 (16.7\%) & 40 (16.7\%) \\
\textbf{F} & 39 (16.2\%) & 62 (25.8\%) & 38 (15.8\%) & 40 (16.7\%) \\
\textbf{N} & 1 (0.4\%) & - & 31 (12.9\%) & 40 (16.7\%) \\
\textbf{M-F-N} & - & - & 103 (42.9\%) & 120 (50\%) \\
\textbf{M-F} & - & 19 (7.9\%) & 3 (1.3\%) & - \\
\textbf{M-N} & - & 1 (0.4\%) & - & - \\
\textbf{error [1]} & 9 (3.7\%) & 3 (1.3\%) & 4 (1.7\%) & - \\
\textbf{error [2]} & 32 (13.3\%) & 3 (1.3\%) & 3 (1.3\%) & - \\
\textbf{error [3]} & - & - & 8 (3.3\%) & - \\
\textbf{error [4]} & 1 (0.4\%) & - & 10 (4.2\%) & - \\ 
\midrule
\textbf{Total} & \textbf{240 (100\%)} & \textbf{240 (100\%)} & \textbf{240 (100\%)} & \textbf{240 (100\%)} \\ 
\bottomrule
\end{tabular}
\caption{Distribution of gender and error labels across the three systems and gold standards, with raw counts and proportions.}
\label{tab:2}
\end{table*}

\subsection{Gender Bias in MT Systems}
\paragraph{Male Bias}

An analysis of the 160 ambiguous English sentences revealed a clear tendency towards male bias. Masculine forms dominated, making up 74.4\% of Google Translate's and 70.6\% of DeepL's outputs. Feminine forms were rare (8.1\% for Google Translate, 13.1\% for DeepL), and errors were higher for Google Translate (17.5\% vs. DeepL's 3.7\%) (\hyperref[sec:appendixD]{Appendix D}).

Most errors by Google Translate were mixed-gender representations all of which appeared in sentences with the singular `they', indicating a difficulty in handling the gender neutrality or non-binarity expressed by the pronoun. Specifically, the model treated `they' as a collective pronoun, defaulting to masculine forms for professions and collective `they' for the pronoun (i.e. ``their work''), failing to correlate the gender-neutral pronoun with gender-neutral solutions in Greek.\footnote{E.g., ``The guard finished their work'' was translated into ``\textgreek{Ο φύλακας τελείωσε τη δουλειά τους}'' (= \textit{The guard [M] finished their [plural] work}).} DeepL's errors, albeit very few, were found in the same `ambiguous + unambiguous
[Non-binary]' sentence type, suggesting that DeepL also struggles with sentences involving neutral or non-binary pronouns.

Regarding gender-inclusive outputs, neither system performed well. Google Translate did not produce any neutral or inclusive translations, while DeepL provided alternatives for 12.5\% of gender-ambiguous sentences, mostly masculine-feminine pairs, with only one case including a neutral variant. These results confirm a \textbf{notable male bias} in both systems, aligning with common findings about male default bias in MT systems.

\paragraph{Occupational Stereotyping}
Masculine translations prevailed in both systems, regardless of stereotype: 82.5\% (Google Translate) and 86.2\% (DeepL) for male-biased occupations, and 66.2\% (Google Translate) and 55\% (DeepL) for female-biased ones. These findings agree with those in the prior section regarding the persistence of male bias in MT systems.

A closer examination of the results reveals an interesting pattern in feminine gender outputs. In the case of male-biased occupations, none of the MT systems produced translations in the feminine form. On the contrary, when the occupation was female-biased, Google Translate generated 16.2\% and DeepL 26.2\% feminine outputs. This signifies that, while masculine remains the default, MT systems are potentially influenced by societal stereotypes, associating feminine forms more frequently with traditionally female-biased professions.

A Fischer's exact test confirmed a significant correlation between occupational stereotypes and gender outputs. The test results for both systems (\hyperref[sec:appendixG]{Appendix G}) verified that the stereotype of the occupation significantly impacts the translation gender, with \textbf{feminine forms more likely to appear for stereotypically female occupations} than stereotypically male ones. Although stereotypes influenced translations, the masculine form remained the overall default.

\paragraph{Anti-Stereotypical Gender Assignments}
Regarding this pattern of bias, the MT systems were expected to generate more frequent errors or incorrect gender assignments when translating anti-stereotypical gender roles, such as ``female doctors'' or ``male nurses'', compared to stereotypical ones, like ``male doctors'' or ``female nurses''. For this pattern of bias, only sentences with feminine and masculine genders were analysed (\hyperref[sec:appendixF]{Appendix F}).

An examination of the \textbf{stereotypical cases} revealed that for male-biased occupations, both systems correctly translated masculine-gendered sentences (`unambiguous [Male]'), where the gender ambiguity was resolved with the use of a masculine pronoun. Only one lexical error by Google Translate was detected: the word ``mover'' was translated as ``\textgreek{μετακινούμενος}'', which refers to someone being ``moved'' rather than the profession of a ``mover''.

For female-biased occupations in female-gendered sentences (`unambiguous [Female]'), DeepL demonstrated consistently accurate performance, correctly assigning the feminine gender to all translations. Google Translate, however, produced three errors (two mixed-gender, one lexical error). Specifically, the system introduced a non-existent feminised form (``\textgreek{φούρναρη}'') as the translation of ``female baker''. This potentially indicates not only a grammatical error but also a difficulty in handling feminine forms for some professions.

Regarding the \textbf{anti-stereotypical cases}, our initial claim is not supported for `unambiguous [Male]' sentences containing female-biased occupations,\footnote{E.g., ``The teacher [F-biased] finished \textit{his} work.''} as Google Translate produced accurate masculine translations in all cases. DeepL also performed well, with only one incorrect gender assignment: ``housekeeper'' was translated into the feminine form.

For `unambiguous [Female]' sentences containing male-biased professions,\footnote{E.g., ``The driver [M-biased] finished \textit{her} work.''} DeepL again showed strong performance, translating all instances into feminine forms. However, Google Translate exhibited notable variability: only 45\% of the sentences were correctly assigned a feminine gender, 50\% were classified as errors (mostly mixed-gender representations), and one instance was labelled as `other'\footnote{Outputs that included masculine-feminine alternatives (M-F), masculine-neutral alternatives (M-N), or exclusively gender-neutral forms (N) were grouped together under the `other' category for illustration purposes.} (translating ``female guard'' as ``\textgreek{η φρουρά}''; neutral term used to refer to the role without specifying the gender). A notable error was translating ``construction worker'' as ``\textgreek{η οικοδομή}'' (the building; lexical error), a noun with a feminine grammatical gender, suggesting that the model, in an attempt to assign feminine form to the output, used an incorrect – yet feminine – term.

Fischer's exact test (\hyperref[sec:appendixG]{Appendix G}) revealed a statistically significant difference for \textbf{Google Translate} in feminine-gendered sentences, showing it \textbf{struggles more when translating male-biased professions into feminine forms}, despite the presence of an explicit feminine pronoun in the source. DeepL outperformed Google Translate, consistently assigning correct genders regardless of stereotypicality. Therefore, the issues in the processing of anti-stereotypical gender assignments remain specific to Google Translate.

\subsection{GPT-4 on Gender Bias Mitigation}

\subsubsection{Quantitative Analysis}
The performance of the prompted GPT-4o showed promising results overall. \hyperref[sec:appendixH]{Appendix H} presents the gender distribution across sentence types. For the \textbf{`unambiguous [Male]'} sentences, the model achieved a 100\% success rate, correctly translating all instances into masculine forms. For \textbf{`unambiguous [Female]'} sentences, it succeeded in 95\% of cases, with 5\% errors.

Furthermore, for the \textbf{`ambiguous base'}, \textbf{`ambiguous + male-biased adj.'}, and \textbf{`ambiguous + female-biased adj.'} sentences, the model showed high precision in detecting gender ambiguity, and generated three alternatives (M-F-N) with success rates of 92.5\%, 80\% and 85\%, respectively. The remaining cases were classified as errors.

Finally, for the \textbf{`ambiguous~/ unambiguous [Non-binary]'} sentences, GPT-4o successfully produced neutral translations using the neutral circumlocution ``\textgreek{το άτομο που}'' (the person who). However, 7.5\% of translations in this category included double forms (e.g. ``\textgreek{ο/η λογιστής/λογίστρια}'' – the [M/F] accountant [M/F]), despite explicit instructions to avoid them. 15\% of translations were classified as errors. 

The above results show that the masculine-gendered sentences obtained the highest accuracy, followed by the feminine-gendered ones. This indicates that the model performed best when the gender of the referent was explicitly specified, with a slightly reduced success rate for feminine forms, possibly reflecting intrinsic bias present in the training data.

A closer comparison of the performance of prompted GPT-4o with Google Translate and DeepL on unambiguous sentences reveals that the LLM outperformed the other two MT systems with 100\% success rate in masculine-gendered sentences (\hyperref[sec:appendixI]{Appendix I}). Interestingly, in \textbf{feminine-gendered sentences}, it was DeepL that had the highest scores (100\%), followed closely by GPT-4o (95\%). Google Translate, in contrast, achieved only 65\% accuracy, with a notable number of incorrect gender assignments and errors. GPT-4o's slight reduction in accuracy for feminine forms may suggest residual biases in its training data.

Overall, GPT-4o handled gender ambiguity effectively, generating three correct alternatives in most cases. However, when (gender-biased) adjectives were included, the rates slightly dropped, indicating that such modifiers introduce additional difficulty. The `ambiguous~/ unambiguous [Non-binary]' sentence type posed the greatest challenge for the model, with the highest percentage of incorrect translations, which may reflect the limitations of the model in producing accurate gender-neutral language. 

\subsubsection{Qualitative Analysis}
A closer investigation of the errors for each sentence category shows interesting information regarding the inaccuracies produced by prompted GPT-4o (\hyperref[sec:appendixH]{Appendix H}). Key observations include:

\paragraph{Ambiguous base} The model consistently produced three gender alternatives for each source sentence, in which the masculine and feminine versions were accurate. However, the neutral forms presented issues: the model attempted to use the neuter circumlocution ``\textgreek{το άτομο που}'' (the person who) but paired with epicene nouns, such as ``\textgreek{μηχανικός}'' (mechanic), terms that maintain binary gender distinctions and fail to accommodate non-binary references. This demonstrates both the model's challenges in achieving true gender neutrality, but also the broader constraints of Greek's available neutralisation strategies.

\paragraph{Ambiguous + male-biased adj.} Once again, all errors occurred in the gender-neutral variants provided by the model, alongside the correct masculine and feminine versions. Most issues in this category, involved the omission of the male-biased adjective from the target sentence. This pattern suggests that the model may struggle to balance its neutralisation efforts with preserving the semantic elements of the source text. In other words, we assume that the model prioritised neutralisation to such an extent that it overlooked key details, such as the male-biased adjective, which is critical for maintaining the original meaning of the sentence.

\paragraph{Ambiguous + female-biased adj.} The model's neutral variants again exhibited three issues: (a) omission of the female-biased adjective in three cases, compromising the original meaning, (b) inaccurate neutralisation attempts using epicenes in two instances, and (c) one syntactically incorrect output featuring redundant repetition of ``\textgreek{που}'' ([the person] who). 

\paragraph{Unambiguous [Female]} The errors here were minimal, with two instances classified as mixed-gender representations. This error shows a mismatch between the grammatical gender of the subject ``sheriff'' and ``farmer'' and the personal pronoun `her' (translated as ``The \textit{male} sheriff/farmer finished \textit{her} work''). While these translations technically align with the source text in terms of pronoun use, the introduction of a masculine article and noun creates a bias and inconsistency that makes the translation somewhat problematic. Instead of fully aligning the output gender-wise based on the feminine pronoun, it defaulted to the masculine form of ``\textgreek{ο σερίφης}'' (the male sheriff) and ``\textgreek{ο αγρότης}'' (the male farmer), possibly influenced by inherent bias of the training data.

\paragraph{Ambiguous / unambiguous [Non-binary]} GPT-4o's most significant challenges emerged in this category, revealing difficulties in producing sufficient gender-neutral translations. First, it frequently defaulted to inadequate solutions, either employing epicenes, such as ``\textgreek{μηχανικός}'' (mechanic) and ``\textgreek{υπάλληλος}'' (clerk), or binary double forms, such as ``\textgreek{ο/η συντάκτης/τρια}'' (the [M/F] editor [M/F]) incorrectly paired with the plural pronoun ``\textgreek{τους}'' (their). This indicates that the model misinterpreted singular `their' as a collective pronoun.

Second, the model generated linguistically invalid forms while attempting neutralisation, including (i) the non-existent ``\textgreek{το ρεσεψιονίστ}'', (ii) repeated feminine possessives in ``\textgreek{τη δουλειά του/της/της}'' (his/her/her work), (iii) the ill-formed ``\textgreek{ο/η/το γραμματέας}'' (the [M/F/N] secretary), suggesting a non-existent neuter form of the specific epicene, and (iv) the completely invented pronoun ``\textgreek{ατους}''. The model apparently struggled to produce gender-neutral language leading to mistranslations and non-existent words.

Third, rather than producing a singular gender-neutral form, GPT-4o often defaulted to listing multiple gendered variations, contradicting prompt instructions for singular `they' translations. In three cases, it generated grammatically correct but non-inclusive outputs, failing to fully adhere to non-binary representations.

\section{Discussion}
Our study confirms significant gender bias in English-to-Greek MT, with both Google Translate and DeepL defaulting to masculine forms for gender-ambiguous contexts, rendering it consistent with findings in other language pairs \citep{prates19, stanovsky-etal-2019-evaluating, currey-etal-2022-mt}. While this male bias persisted across occupational stereotypes, increased feminine forms for stereotypically female occupations demonstrate how systems simultaneously reinforce male defaults and societal gender associations embedded in the systems' training data \citep{savoldi-etal-2021-gender}.

Both MT systems performed consistently well with explicitly masculine referents, supporting previous work showing better handling of male references and stereotypical roles \citep{sun-etal-2019-mitigating, kocmi-etal-2020-gender, saundersbill20}. However, Google Translate struggled more with anti-stereotypical feminine forms. This disparity is likely influenced by differences in training data or model architecture, but the black-box nature of these systems makes it difficult to determine the exact cause. Most critically, neither system produced (adequate) gender-neutral outputs, consistently failing on non-binary cases.

Prompted GPT-4o demonstrated high performance, successfully generating feminine, neutral and masculine alternatives for most ambiguous cases. While it showed residual bias (better accuracy in masculine forms), the model was able to follow instructions to identify gender-ambiguous sentences and generate inclusive output for them. We thus confirmed LLMs' potential to handle gender-aware translation, supporting emerging research \citep{savoldi24-prompt} that also demonstrates that GPT is a promising solution for producing gender-neutral outputs when given only a few examples.

Nevertheless, the translation errors (e.g. non-existent words, incorrect pronouns, missing adjectives) or incorrect neutralisation solutions (e.g. use of epicene nouns, double forms with binary pronouns) that were produced for a small number of gender-neutral outputs, should be taken into consideration. These issues underscore the tension between technological solutions and linguistic reality; that is, an inherent bias in the training data of the model, but also the challenges in adapting gender-neutral practices for Greek. Language reform in Greek has followed a slower pace compared to other languages and currently lacks sufficient linguistic structures for a gender-neutral language or structures that address the visibility of under-represented groups, such as LGBTQIA+ individuals and women. As such, the discussion about linguistic sexism and the development of gender-inclusive practices is still open and evolving.

\section{Limitations}

Some limitations should be acknowledged. First, there is a reproducibility problem as the study relies on three closed-source models. As proprietary systems subject to frequent updates, the results of the same query may vary across multiple trials. Moreover, GPT-4o requires a paid subscription,\footnote{\$2.50/1M input tokens and \$10.00/1M output tokens} which limits its accessibility compared to freely available systems.

Second, while GendEL's controlled sentence structures (based on gender-biased adjectives and occupational nouns) ensure methodological consistency, this design restricts the generalisability of the results to more diverse and natural text. Similarly, our GPT-4o prompt was specifically customised for these sentence structures, which raises questions about its applicability to more complex data. GendEL, therefore, should be viewed as a foundational resource for evaluating English-to-Greek gender bias, and future research could supplement a wider variety of sentence structures, contexts, linguistic phenomena, and manifestations of gender bias, as well as further experimentation with LLM prompting strategies.

\section{Conclusion}

In response to the emerging demand for inclusive language, this study focused on the under-represented English-to-Greek language pair. Through extensive, fine-grained manual analyses and descriptive statistics, we demonstrated that gender bias is persistent in translations by Google Translate and DeepL, highlighting that, while they perform well in cases where the referent's gender is defined, they are far from recognising and producing gender-neutral language. We also demonstrated that GPT-4o, when prompted, can achieve high accuracy on providing gendered and gender-neutral alternatives in cases of ambiguity. By situating our results within the context of prior research, this study makes two important contributions: (1) the creation and public release of GendEL, the first handcrafted dataset for evaluating English-to-Greek translations, and (2) empirical evidence emphasising the urgent need for more inclusive translation practices in Greek. We hope this work will inspire further research on this language pair and contribute to the development of more inclusive translation technologies.

\bibliography{mtsummit25}

\begin{thebibliography}{58}
\providecommand{\natexlab}[1]{#1}

\bibitem[{Alvanoudi(2015)}]{alvanoudi15book}
Angeliki Alvanoudi. 2015.
\newblock \href {https://doi.org/10.1163/9789004283152} {\emph{Grammatical gender in interaction: Cultural and cognitive aspects}}.
\newblock Brill.

\bibitem[{Asscher and Glikson(2023)}]{asscher_human_2023}
Omri Asscher and Ella Glikson. 2023.
\newblock \href {https://doi.org/10.1177/14614448211018833} {Human evaluations of machine translation in an ethically charged situation}.
\newblock \emph{New Media \& Society}, 25(5):1087--1107.
\newblock Publisher: SAGE Publications.

\bibitem[{Blodgett et~al.(2020)Blodgett, Barocas, Daumé~III, and Wallach}]{blodgett_20}
Su~Lin Blodgett, Solon Barocas, Hal Daumé~III, and Hanna Wallach. 2020.
\newblock \href {https://doi.org/10.18653/v1/2020.acl-main.485} {Language ({Technology}) is {Power}: {A} {Critical} {Survey} of “{Bias}” in {NLP}}.
\newblock In \emph{Proceedings of the 58th {Annual} {Meeting} of the {Association} for {Computational} {Linguistics}}, pages 5454--5476, Online. Association for Computational Linguistics.

\bibitem[{Bolukbasi et~al.(2016)Bolukbasi, Chang, Zou, Saligrama, and Kalai}]{bolukbasi16}
Tolga Bolukbasi, Kai-Wei Chang, James Zou, Venkatesh Saligrama, and Adam Kalai. 2016.
\newblock \href {https://arxiv.org/abs/1607.06520} {Man is to computer programmer as woman is to homemaker? debiasing word embeddings}.
\newblock \emph{Preprint}, arXiv:1607.06520.

\bibitem[{Cho et~al.(2019)Cho, Kim, Kim, and Kim}]{cho19}
Won~Ik Cho, Ji~Won Kim, Seok~Min Kim, and Nam~Soo Kim. 2019.
\newblock \href {https://doi.org/10.18653/v1/W19-3824} {On measuring gender bias in translation of gender-neutral pronouns}.
\newblock In \emph{Proceedings of the First Workshop on Gender Bias in Natural Language Processing}, pages 173--181, Florence, Italy. Association for Computational Linguistics.

\bibitem[{Currey et~al.(2022)Currey, Nadejde, Pappagari, Mayer, Lauly, Niu, Hsu, and Dinu}]{currey-etal-2022-mt}
Anna Currey, Maria Nadejde, Raghavendra~Reddy Pappagari, Mia Mayer, Stanislas Lauly, Xing Niu, Benjamin Hsu, and Georgiana Dinu. 2022.
\newblock \href {https://doi.org/10.18653/v1/2022.emnlp-main.288} {{MT}-{G}en{E}val: A counterfactual and contextual dataset for evaluating gender accuracy in machine translation}.
\newblock In \emph{Proceedings of the 2022 Conference on Empirical Methods in Natural Language Processing}, pages 4287--4299, Abu Dhabi, United Arab Emirates. Association for Computational Linguistics.

\bibitem[{Elaraby et~al.(2018)Elaraby, Tawfik, Khaled, Hassan, and Osama}]{elaraby18}
Mostafa Elaraby, Ahmed~Y. Tawfik, Mahmoud Khaled, Hany Hassan, and Aly Osama. 2018.
\newblock \href {https://doi.org/10.1109/ICNLSP.2018.8374387} {Gender aware spoken language translation applied to english-arabic}.
\newblock In \emph{2018 2nd International Conference on Natural Language and Speech Processing (ICNLSP)}, pages 1--6.

\bibitem[{Escud{\'e}~Font and Costa-juss{\`a}(2019)}]{escude19}
Joel Escud{\'e}~Font and Marta~R. Costa-juss{\`a}. 2019.
\newblock \href {https://doi.org/10.18653/v1/W19-3821} {Equalizing gender bias in neural machine translation with word embeddings techniques}.
\newblock In \emph{Proceedings of the First Workshop on Gender Bias in Natural Language Processing}, pages 147--154, Florence, Italy. Association for Computational Linguistics.

\bibitem[{Fisher(1992)}]{fisher_92}
R.~A. Fisher. 1992.
\newblock \href {https://doi.org/10.1007/978-1-4612-4380-9_6} {\emph{Statistical Methods for Research Workers}}, pages 66--70.
\newblock Springer New York, New York, NY.

\bibitem[{Freitag et~al.(2021)Freitag, Foster, Grangier, Ratnakar, Tan, and Macherey}]{freitag_experts_2021}
Markus Freitag, George Foster, David Grangier, Viresh Ratnakar, Qijun Tan, and Wolfgang Macherey. 2021.
\newblock \href {https://doi.org/10.1162/tacl_a_00437} {Experts, {Errors}, and {Context}: {A} {Large}-{Scale} {Study} of {Human} {Evaluation} for {Machine} {Translation}}.
\newblock \emph{Transactions of the Association for Computational Linguistics}, 9:1460--1474.
\newblock Place: Cambridge, MA Publisher: MIT Press.

\bibitem[{Friedman and Nissenbaum(1996)}]{friedman_1996}
Batya Friedman and Helen Nissenbaum. 1996.
\newblock \href {https://doi.org/10.1145/230538.230561} {Bias in computer systems}.
\newblock \emph{ACM Trans. Inf. Syst.}, 14(3):330--347.

\bibitem[{Ghosh and Caliskan(2023)}]{ghosh23}
Sourojit Ghosh and Aylin Caliskan. 2023.
\newblock \href {https://doi.org/10.1145/3600211.3604672} {Chatgpt perpetuates gender bias in machine translation and ignores non-gendered pronouns: Findings across bengali and five other low-resource languages}.
\newblock In \emph{Proceedings of the 2023 AAAI/ACM Conference on AI, Ethics, and Society}, AIES '23, page 901–912, New York, NY, USA. Association for Computing Machinery.

\bibitem[{Gonen and Webster(2020)}]{gonen20}
Hila Gonen and Kellie Webster. 2020.
\newblock \href {https://doi.org/10.18653/v1/2020.findings-emnlp.180} {Automatically identifying gender issues in machine translation using perturbations}.
\newblock In \emph{Findings of the Association for Computational Linguistics: EMNLP 2020}, pages 1991--1995, Online. Association for Computational Linguistics.

\bibitem[{Johnson(2020)}]{johnson20}
Melvin Johnson. 2020.
\newblock \href {https://ai.googleblog.com/2020/40/04/a-scalable-approach-to-reducing-gender.html} {A scalable approach to reducing gender bias in google translate}.

\bibitem[{Kalfadopoulou and Tsigou(2022)}]{kalfadopoulou22}
Valentini Kalfadopoulou and Maria Tsigou. 2022.
\newblock Inclusive language in translation technology: Theory and practice; the case of greek.
\newblock In \emph{Proceedings of the New Trends in Translation and Technology Conference - NeTTT 2022}, pages 206--213, Rhodes Island, Greece.

\bibitem[{Karastergiou and Diamantopoulos(2024)}]{karastergiou24}
Anestis~Polychronis Karastergiou and Konstantinos Diamantopoulos. 2024.
\newblock Gender issues in machine translation.
\newblock \emph{Transcultural Journal of Humanities \& Social Sciences}, 5:48--64.

\bibitem[{Kocmi et~al.(2020)Kocmi, Limisiewicz, and Stanovsky}]{kocmi-etal-2020-gender}
Tom Kocmi, Tomasz Limisiewicz, and Gabriel Stanovsky. 2020.
\newblock \href {https://aclanthology.org/2020.wmt-1.39/} {Gender coreference and bias evaluation at {WMT} 2020}.
\newblock In \emph{Proceedings of the Fifth Conference on Machine Translation}, pages 357--364, Online. Association for Computational Linguistics.

\bibitem[{Lardelli et~al.(2024)Lardelli, Attanasio, and Lauscher}]{Lardelli:24}
Manuel Lardelli, Giuseppe Attanasio, and Anne Lauscher. 2024.
\newblock \href {https://doi.org/10.18653/v1/2024.findings-acl.448} {Building bridges: A dataset for evaluating gender-fair machine translation into {G}erman}.
\newblock In \emph{Findings of the Association for Computational Linguistics: ACL 2024}, pages 7542--7550, Bangkok, Thailand. Association for Computational Linguistics.

\bibitem[{Lardelli and Gromann(2023)}]{lardelli_23}
Manuel Lardelli and Dagmar Gromann. 2023.
\newblock Translating non-binary coming-out reports: {Gender}-fair language strategies and use in news articles.
\newblock \emph{The Journal of Specialised Translation}, pages 213--240.

\bibitem[{Lee et~al.(2024)Lee, Koh, Kim, and Jung}]{lee24}
Minwoo Lee, Hyukhun Koh, Minsung Kim, and Kyomin Jung. 2024.
\newblock \href {https://doi.org/10.18653/v1/2024.naacl-long.303} {Fine-grained gender control in machine translation with large language models}.
\newblock In \emph{Proceedings of the 2024 Conference of the North American Chapter of the Association for Computational Linguistics: Human Language Technologies (Volume 1: Long Papers)}, pages 5416--5430, Mexico City, Mexico. Association for Computational Linguistics.

\bibitem[{Levy et~al.(2021)Levy, Lazar, and Stanovsky}]{levy21}
Shahar Levy, Koren Lazar, and Gabriel Stanovsky. 2021.
\newblock \href {https://doi.org/10.18653/v1/2021.findings-emnlp.211} {Collecting a large-scale gender bias dataset for coreference resolution and machine translation}.
\newblock In \emph{Findings of the Association for Computational Linguistics: EMNLP 2021}, pages 2470--2480, Punta Cana, Dominican Republic. Association for Computational Linguistics.

\bibitem[{Mastromichalakis et~al.(2024)Mastromichalakis, Filandrianos, Tsouparopoulou, Parsanoglou, Symeonaki, and Stamou}]{Mastromichalakis:24}
Orfeas~Menis Mastromichalakis, Giorgos Filandrianos, Eva Tsouparopoulou, Dimitris Parsanoglou, Maria Symeonaki, and Giorgos Stamou. 2024.
\newblock \href {https://arxiv.org/abs/2409.10989} {Gost-mt: A knowledge graph for occupation-related gender biases in machine translation}.
\newblock \emph{Preprint}, arXiv:2409.10989.

\bibitem[{M{\v{e}}chura(2022)}]{mechura22}
Michal M{\v{e}}chura. 2022.
\newblock \href {https://doi.org/10.18653/v1/2022.gebnlp-1.18} {A taxonomy of bias-causing ambiguities in machine translation}.
\newblock In \emph{Proceedings of the 4th Workshop on Gender Bias in Natural Language Processing (GeBNLP)}, pages 168--173, Seattle, Washington. Association for Computational Linguistics.

\bibitem[{Monti(2020)}]{monti_20}
Johanna Monti. 2020.
\newblock Gender issues in machine translation: {An} unsolved problem?
\newblock In \emph{The {Routledge} {Handbook} of {Translation}, {Feminism} and {Gender}}. Routledge.

\bibitem[{Mucchi-Faina(2005)}]{mucchi05}
Angelica Mucchi-Faina. 2005.
\newblock \href {https://doi.org/10.1177/0539018405050466} {{Visible or influential? Language reforms and gender (in)equality}}.
\newblock \emph{Social Science Information}, 44(1):189--215.

\bibitem[{Ntouvlis(2020)}]{ntouvlis20}
Vinicio Ntouvlis. 2020.
\newblock Online writing and linguistic sexism: The use of gender-inclusive @ on a greek feminist facebook page.
\newblock \emph{Tilburg Papers in Culture Studies}, 245.

\bibitem[{Nurminen and Koponen(2020)}]{nurminen_machine_2020}
Mary Nurminen and Maarit Koponen. 2020.
\newblock \href {https://doi.org/10.1075/ts.00025.nur} {Machine translation and fair access to information}.
\newblock \emph{Translation Spaces}, 9(1):150--169.
\newblock Publisher: John Benjamins Publishing Company.

\bibitem[{Papineni et~al.(2002)Papineni, Roukos, Ward, and Zhu}]{papineni02}
Kishore Papineni, Salim Roukos, Todd Ward, and Wei-Jing Zhu. 2002.
\newblock \href {https://doi.org/10.3115/1073083.1073135} {{B}leu: a method for automatic evaluation of machine translation}.
\newblock In \emph{Proceedings of the 40th Annual Meeting of the Association for Computational Linguistics}, pages 311--318, Philadelphia, Pennsylvania, USA. Association for Computational Linguistics.

\bibitem[{Pavlidou et~al.(2004)Pavlidou, Alvanoudi, and Karafoti}]{pavlidou_grammatical_2004}
Theodossia-Soula Pavlidou, Angeliki Alvanoudi, and Eleni Karafoti. 2004.
\newblock Grammatical gender and semantic content: preliminary remarks on the lexical representation of social gender [in {Greek}].
\newblock In \emph{Studies in {Greek} {Linguistics}}, volume~24, pages 543--553. Aristotle University of Thessaloniki.

\bibitem[{Piergentili et~al.(2023)Piergentili, Fucci, Savoldi, Bentivogli, and Negri}]{piergentili23}
Andrea Piergentili, Dennis Fucci, Beatrice Savoldi, Luisa Bentivogli, and Matteo Negri. 2023.
\newblock \href {https://aclanthology.org/2023.gitt-1.7/} {Gender neutralization for an inclusive machine translation: from theoretical foundations to open challenges}.
\newblock In \emph{Proceedings of the First Workshop on Gender-Inclusive Translation Technologies}, pages 71--83, Tampere, Finland. European Association for Machine Translation.

\bibitem[{Piergentili et~al.(2024)Piergentili, Savoldi, Negri, and Bentivogli}]{piergentili24}
Andrea Piergentili, Beatrice Savoldi, Matteo Negri, and Luisa Bentivogli. 2024.
\newblock \href {https://aclanthology.org/2024.eamt-1.25} {Enhancing gender-inclusive machine translation with neomorphemes and large language models}.
\newblock In \emph{Proceedings of the 25th Annual Conference of the European Association for Machine Translation (Volume 1)}, pages 300--314, Sheffield, UK. European Association for Machine Translation (EAMT).

\bibitem[{Prates et~al.(2019)Prates, Avelar, and Lamb}]{prates19}
Marcelo O.~R. Prates, Pedro H.~C. Avelar, and Luis Lamb. 2019.
\newblock \href {https://arxiv.org/abs/1809.02208} {Assessing gender bias in machine translation -- a case study with google translate}.
\newblock \emph{Preprint}, arXiv:1809.02208.

\bibitem[{Rarrick et~al.(2024)Rarrick, Naik, Poudel, and Chowdhary}]{rarrick24}
Spencer Rarrick, Ranjita Naik, Sundar Poudel, and Vishal Chowdhary. 2024.
\newblock \href {https://arxiv.org/abs/2402.14277} {Gate x-e : A challenge set for gender-fair translations from weakly-gendered languages}.
\newblock \emph{Preprint}, arXiv:2402.14277.

\bibitem[{Robinson et~al.(2024)Robinson, Kudugunta, Stella, Dev, and Bastings}]{robinson24}
Kevin Robinson, Sneha Kudugunta, Romina Stella, Sunipa Dev, and Jasmijn Bastings. 2024.
\newblock \href {https://arxiv.org/abs/2401.06935} {Mittens: A dataset for evaluating gender mistranslation}.
\newblock \emph{Preprint}, arXiv:2401.06935.

\bibitem[{Rudinger et~al.(2018)Rudinger, Naradowsky, Leonard, and Van~Durme}]{rudinger18}
Rachel Rudinger, Jason Naradowsky, Brian Leonard, and Benjamin Van~Durme. 2018.
\newblock \href {https://doi.org/10.18653/v1/N18-2002} {Gender bias in coreference resolution}.
\newblock In \emph{Proceedings of the 2018 Conference of the North {A}merican Chapter of the Association for Computational Linguistics: Human Language Technologies, Volume 2 (Short Papers)}, pages 8--14, New Orleans, Louisiana. Association for Computational Linguistics.

\bibitem[{S{\'a}nchez et~al.(2024)S{\'a}nchez, Andrews, Stenetorp, Artetxe, and Costa-juss{\`a}}]{sanchez24}
Eduardo S{\'a}nchez, Pierre Andrews, Pontus Stenetorp, Mikel Artetxe, and Marta~R. Costa-juss{\`a}. 2024.
\newblock \href {https://aclanthology.org/2024.mrl-1.10} {Gender-specific machine translation with large language models}.
\newblock In \emph{Proceedings of the Fourth Workshop on Multilingual Representation Learning (MRL 2024)}, pages 148--158, Miami, Florida, USA. Association for Computational Linguistics.

\bibitem[{Saunders and Byrne(2020)}]{saundersbill20}
Danielle Saunders and Bill Byrne. 2020.
\newblock \href {https://doi.org/10.18653/v1/2020.acl-main.690} {Reducing gender bias in neural machine translation as a domain adaptation problem}.
\newblock In \emph{Proceedings of the 58th Annual Meeting of the Association for Computational Linguistics}, pages 7724--7736, Online. Association for Computational Linguistics.

\bibitem[{Saunders et~al.(2020)Saunders, Sallis, and Byrne}]{saunders20}
Danielle Saunders, Rosie Sallis, and Bill Byrne. 2020.
\newblock \href {https://aclanthology.org/2020.gebnlp-1.4} {Neural machine translation doesn{'}t translate gender coreference right unless you make it}.
\newblock In \emph{Proceedings of the Second Workshop on Gender Bias in Natural Language Processing}, pages 35--43, Barcelona, Spain (Online). Association for Computational Linguistics.

\bibitem[{Savoldi et~al.(2021)Savoldi, Gaido, Bentivogli, Negri, and Turchi}]{savoldi-etal-2021-gender}
Beatrice Savoldi, Marco Gaido, Luisa Bentivogli, Matteo Negri, and Marco Turchi. 2021.
\newblock \href {https://doi.org/10.1162/tacl_a_00401} {Gender bias in machine translation}.
\newblock \emph{Transactions of the Association for Computational Linguistics}, 9:845--874.

\bibitem[{Savoldi et~al.(2024)Savoldi, Piergentili, Fucci, Negri, and Bentivogli}]{savoldi24-prompt}
Beatrice Savoldi, Andrea Piergentili, Dennis Fucci, Matteo Negri, and Luisa Bentivogli. 2024.
\newblock \href {https://aclanthology.org/2024.eacl-short.23} {A prompt response to the demand for automatic gender-neutral translation}.
\newblock In \emph{Proceedings of the 18th Conference of the European Chapter of the Association for Computational Linguistics (Volume 2: Short Papers)}, pages 256--267, St. Julian{'}s, Malta. Association for Computational Linguistics.

\bibitem[{Sennrich(2017)}]{sennrich17}
Rico Sennrich. 2017.
\newblock \href {https://aclanthology.org/E17-2060} {How grammatical is character-level neural machine translation? assessing {MT} quality with contrastive translation pairs}.
\newblock In \emph{Proceedings of the 15th Conference of the {E}uropean Chapter of the Association for Computational Linguistics: Volume 2, Short Papers}, pages 376--382, Valencia, Spain. Association for Computational Linguistics.

\bibitem[{Snover et~al.(2006)Snover, Dorr, Schwartz, Micciulla, and Makhoul}]{snover06}
Matthew Snover, Bonnie Dorr, Rich Schwartz, Linnea Micciulla, and John Makhoul. 2006.
\newblock \href {https://aclanthology.org/2006.amta-papers.25} {A study of translation edit rate with targeted human annotation}.
\newblock In \emph{Proceedings of the 7th Conference of the Association for Machine Translation in the Americas: Technical Papers}, pages 223--231, Cambridge, Massachusetts, USA. Association for Machine Translation in the Americas.

\bibitem[{Stafanovi{\v{c}}s et~al.(2020)Stafanovi{\v{c}}s, Bergmanis, and Pinnis}]{stafanovics20}
Art{\=u}rs Stafanovi{\v{c}}s, Toms Bergmanis, and M{\=a}rcis Pinnis. 2020.
\newblock \href {https://aclanthology.org/2020.wmt-1.73} {Mitigating gender bias in machine translation with target gender annotations}.
\newblock In \emph{Proceedings of the Fifth Conference on Machine Translation}, pages 629--638, Online. Association for Computational Linguistics.

\bibitem[{Stanovsky et~al.(2019)Stanovsky, Smith, and Zettlemoyer}]{stanovsky-etal-2019-evaluating}
Gabriel Stanovsky, Noah~A. Smith, and Luke Zettlemoyer. 2019.
\newblock \href {https://doi.org/10.18653/v1/P19-1164} {Evaluating gender bias in machine translation}.
\newblock In \emph{Proceedings of the 57th Annual Meeting of the Association for Computational Linguistics}, pages 1679--1684, Florence, Italy. Association for Computational Linguistics.

\bibitem[{Sun et~al.(2019)Sun, Gaut, Tang, Huang, ElSherief, Zhao, Mirza, Belding, Chang, and Wang}]{sun-etal-2019-mitigating}
Tony Sun, Andrew Gaut, Shirlyn Tang, Yuxin Huang, Mai ElSherief, Jieyu Zhao, Diba Mirza, Elizabeth Belding, Kai-Wei Chang, and William~Yang Wang. 2019.
\newblock \href {https://doi.org/10.18653/v1/P19-1159} {Mitigating gender bias in natural language processing: Literature review}.
\newblock In \emph{Proceedings of the 57th Annual Meeting of the Association for Computational Linguistics}, pages 1630--1640, Florence, Italy. Association for Computational Linguistics.

\bibitem[{\textgreek{Γκασούκα} and \textgreek{Γεωργαλίδου}(2018)}]{gkasouka2018}
\textgreek{Μαρία} \textgreek{Γκασούκα} and \textgreek{Μαριάνθη} \textgreek{Γεωργαλίδου}. 2018.
\newblock \textgreek{Οδηγός μη σεξιστικής γλώσσας στα διοικητικά έγγραφα}.

\bibitem[{\textgreek{Σαρρή-Χασάν}(2024)}]{sarri24}
\textgreek{Ντενίζ} \textgreek{Σαρρή-Χασάν}. 2024.
\newblock \emph{\textgreek{Πρακτικός Οδηγός για τη Χρήση Συμπεριληπτικής ως προς το Φύλο Γλώσσας στο ΕΑΠ}}.
\newblock \textgreek{Εκδόσεις ΕΑΠ Α.Ε.}

\bibitem[{\textgreek{Τριανταφυλλίδης}(1963)}]{triantaf63}
\textgreek{Μανόλης} \textgreek{Τριανταφυλλίδης}. 1963.
\newblock \emph{\textgreek{Η βουλευτίνα και ο σχηματισμός των θηλυκών επαγγελματικών}}, volume \textgreek{Β}, pages 326--334.
\newblock \textgreek{Ίδρυμα Μανόλη Τριανταφυλλίδη}.

\bibitem[{\textgreek{Τσοκαλίδου}(1996)}]{tsokalidou96}
\textgreek{Ρούλα} \textgreek{Τσοκαλίδου}. 1996.
\newblock \emph{\textgreek{To Φύλο της Γλώσσας, Oδηγός μη-σεξιστικής γλώσσας για τον δημόσιο ελληνικό λόγο}}.
\newblock \textgreek{Σύνδεσμος Ελληνίδων Επιστημόνων-Βιβλιοπωλείο της Εστίας}.

\bibitem[{\textgreek{Τσοπανάκης}(1982)}]{tsopanakis82}
\textgreek{Αγαπητός} \textgreek{Τσοπανάκης}. 1982.
\newblock \textgreek{Ο δρόμος προς την δημοτική: Θεωρητικά, τεχνικά και γλωσσικά προβλήματα. Σχηματισμός επαγγελματικών θηλυκών}.
\newblock In \emph{\textgreek{Ὁ δρόμος πρός τήν Δημοτική (Μελέτες και ἄρθρα)}}, pages 302--342. \textgreek{Εκδοτικός οίκος Αφών Κυριακίδη}.

\bibitem[{Troles and Schmid(2021)}]{troles21}
Jonas-Dario Troles and Ute Schmid. 2021.
\newblock \href {https://aclanthology.org/2021.wmt-1.61} {Extending challenge sets to uncover gender bias in machine translation: Impact of stereotypical verbs and adjectives}.
\newblock In \emph{Proceedings of the Sixth Conference on Machine Translation}, pages 531--541, Online. Association for Computational Linguistics.

\bibitem[{Vanmassenhove(2024)}]{Vanmassenhove24}
Eva Vanmassenhove. 2024.
\newblock \href {https://api.semanticscholar.org/CorpusID:267034972} {Gender bias in machine translation and the era of large language models}.
\newblock \emph{ArXiv}, abs/2401.10016.

\bibitem[{Vanmassenhove et~al.(2021)Vanmassenhove, Emmery, and Shterionov}]{vanmassenhove-etal-2021-neutral}
Eva Vanmassenhove, Chris Emmery, and Dimitar Shterionov. 2021.
\newblock \href {https://doi.org/10.18653/v1/2021.emnlp-main.704} {{N}eu{T}ral {R}ewriter: {A} rule-based and neural approach to automatic rewriting into gender neutral alternatives}.
\newblock In \emph{Proceedings of the 2021 Conference on Empirical Methods in Natural Language Processing}, pages 8940--8948, Online and Punta Cana, Dominican Republic. Association for Computational Linguistics.

\bibitem[{Vanmassenhove et~al.(2018)Vanmassenhove, Hardmeier, and Way}]{vanmassenhove-etal-2018-getting}
Eva Vanmassenhove, Christian Hardmeier, and Andy Way. 2018.
\newblock \href {https://doi.org/10.18653/v1/D18-1334} {Getting gender right in neural machine translation}.
\newblock In \emph{Proceedings of the 2018 Conference on Empirical Methods in Natural Language Processing}, pages 3003--3008, Brussels, Belgium. Association for Computational Linguistics.

\bibitem[{Vieira(2020)}]{vieira_machine_2020}
Lucas~Nunes Vieira. 2020.
\newblock \href {https://doi.org/10.1075/ts.00023.nun} {Machine translation in the news: {A} framing analysis of the written press}.
\newblock \emph{Translation Spaces}, 9(1):98--122.
\newblock Publisher: John Benjamins.

\bibitem[{Zhao et~al.(2018)Zhao, Wang, Yatskar, Ordonez, and Chang}]{zhao18}
Jieyu Zhao, Tianlu Wang, Mark Yatskar, Vicente Ordonez, and Kai-Wei Chang. 2018.
\newblock \href {https://doi.org/10.18653/v1/N18-2003} {Gender bias in coreference resolution: Evaluation and debiasing methods}.
\newblock In \emph{Proceedings of the 2018 Conference of the North {A}merican Chapter of the Association for Computational Linguistics: Human Language Technologies, Volume 2 (Short Papers)}, pages 15--20, New Orleans, Louisiana. Association for Computational Linguistics.

\bibitem[{Zhao et~al.(2024)Zhao, Ding, Jia, Wang, and Qian}]{Zhao:24}
Jinman Zhao, Yitian Ding, Chen Jia, Yining Wang, and Zifan Qian. 2024.
\newblock \href {https://arxiv.org/abs/2403.00277} {Gender bias in large language models across multiple languages}.
\newblock \emph{Preprint}, arXiv:2403.00277.

\bibitem[{Zmigrod et~al.(2019)Zmigrod, Mielke, Wallach, and Cotterell}]{zmigrod19}
Ran Zmigrod, Sabrina~J. Mielke, Hanna Wallach, and Ryan Cotterell. 2019.
\newblock \href {https://doi.org/10.18653/v1/P19-1161} {Counterfactual data augmentation for mitigating gender stereotypes in languages with rich morphology}.
\newblock In \emph{Proceedings of the 57th Annual Meeting of the Association for Computational Linguistics}, pages 1651--1661, Florence, Italy. Association for Computational Linguistics.

\end{thebibliography}

\pagebreak

\onecolumn 
\appendix

\section{Stereotypical Occupations and Adjectives}
\label{sec:appendixA}

\begin{table}[htbp]
\centering
\begin{tabular}{@{}ll@{}}
\toprule
\textbf{Male-biased occupations} & \begin{tabular}[c]{@{}l@{}}carpenter (3\%), construction worker (4\%), laborer (4\%), \\ mechanic (4\%), driver (20\%), mover (18\%), sheriff (18\%), \\ developer (20\%), guard (22\%), farmer (25\%), chief (28\%), \\ lawyer (36\%), janitor (37\%), CEO (39\%), analyst (41\%), \\ physician (41\%), cook (42\%), manager (43\%), supervisor (44\%), \\ salesperson (48\%) \end{tabular} \\ \midrule
\textbf{Female-biased occupations} & \begin{tabular}[c]{@{}l@{}}designer (54\%), baker (60\%), accountant (62\%), auditor (62\%), \\ editor (63\%), writer (63\%), cashier (71\%), clerk (72\%), \\ tailor (75\%), attendant (76\%), counselor (76\%), teacher (78\%), \\ librarian (80\%), assistant (85\%), cleaner (89\%), housekeeper (89\%), \\ receptionist (89\%), nurse (90\%), hairdresser (92\%), secretary (93\%)\end{tabular} \\ \bottomrule
\end{tabular}
\caption{Male-biased and female-biased occupations included in GendEL. The percentage of women in the occupation in the US is displayed in brackets \citep{troles21}.}
\end{table}

\begin{table}[htbp]
\centering
\begin{tabular}{@{}ll@{}}
\toprule
\textbf{Male-biased adjectives} & \begin{tabular}[c]{@{}l@{}}grizzled, affable, jovial, suave, debonair, wiry, rascally, \\ arrogant, shifty, eminent\end{tabular} \\ \midrule
\textbf{Female-biased adjectives} & \begin{tabular}[c]{@{}l@{}}sassy, perky, brunette, blonde, lovely, vivacious, saucy, \\ bubbly, alluring, married\end{tabular} \\ \bottomrule
\end{tabular}
\caption{Male-biased and female-biased adjectives included in GendEL.}
\label{tab:3}
\end{table}

\section{GendEL Sample}
\label{sec:appendixB}

\begin{figure}[htbp]
  \centering
  \fbox{\includegraphics[width=1\textwidth]{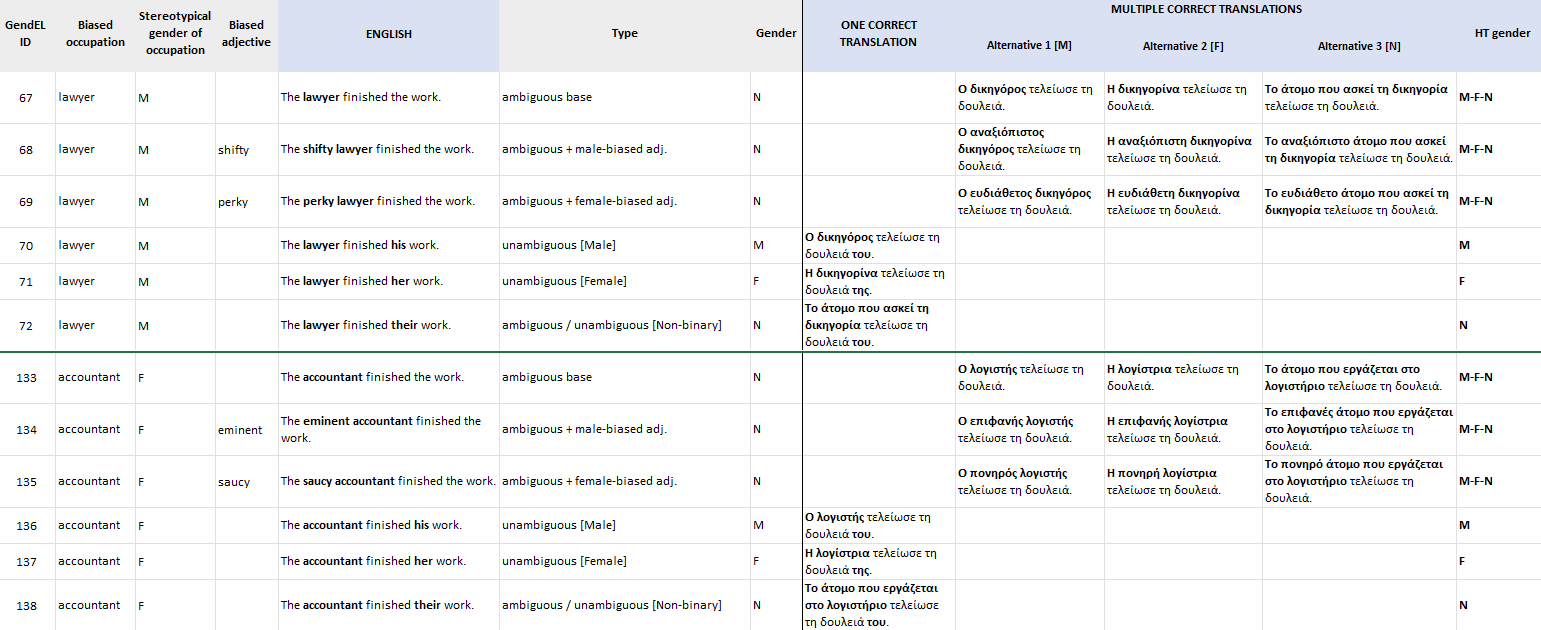}}
  \caption{Subset of a male-biased (``lawyer'') and a subset of a female-biased (``accountant'') occupation.}
\end{figure}

\section{Prompt for GPT-4o}
\label{sec:appendixC}

You are a machine translation assistant focused on gender-fair translations. Translate the given English text into Greek following these rules:

\begin{enumerate}
    \item If the gender of the referent is defined, translate according to that gender. Only ONE translation is correct. E.g.:

\textbf{Input:} The student finished his work. \\
\textbf{Translation:} \textgreek{Ο μαθητής τελείωσε τη δουλειά του}.

\item If the gender of the referent is not defined, provide three alternatives: masculine, feminine, neutral. E.g.: \\
\textbf{Input:} The happy professor finished the work. \\
\textbf{Translations:} \\
a) Male: \textgreek{Ο χαρούμενος καθηγητής τελείωσε τη δουλειά}. \\
b) Female: \textgreek{Η χαρούμενη καθηγήτρια τελείωσε τη δουλειά}. \\
c) Neutral: \textgreek{Το χαρούμενο μέλος του εκπαιδευτικού προσωπικού τελείωσε τη δουλειά}.

\textbf{Input:} The inspector finished the work. \\
\textbf{Translations:} \\
a) Male: \textgreek{Ο επιθεωρητής τελείωσε τη δουλειά}. \\
b) Female: \textgreek{Η επιθεωρήτρια τελείωσε τη δουλειά}. \\
c) Neutral: \textgreek{Το άτομο που επιθεωρεί τελείωσε τη δουλειά}. 

\end{enumerate}

\noindent \textbf{IMPORTANT:} Pay attention to identifying the non-binary singular `they' pronoun, which is used for non-binary individuals. If spotted, you must provide only the neutral version, e.g. ``\textgreek{Το άτομο που επιθεωρεί τελείωσε τη δουλειά του}''. \\

\noindent \textbf{IMPORTANT:} For the neutral version, do NOT indicate any gender. Avoid forms like ``\textgreek{ο επιθεωρητής}'' or ``\textgreek{ο/η επιθεωρητής/τρια}''. \\

\noindent Translate this text: \{input\_text\}

\section{Gender Bias in MT Systems: Male Bias}
\label{sec:appendixD}

\begin{figure}[htbp]
  \centering
  \includegraphics[width=\columnwidth]{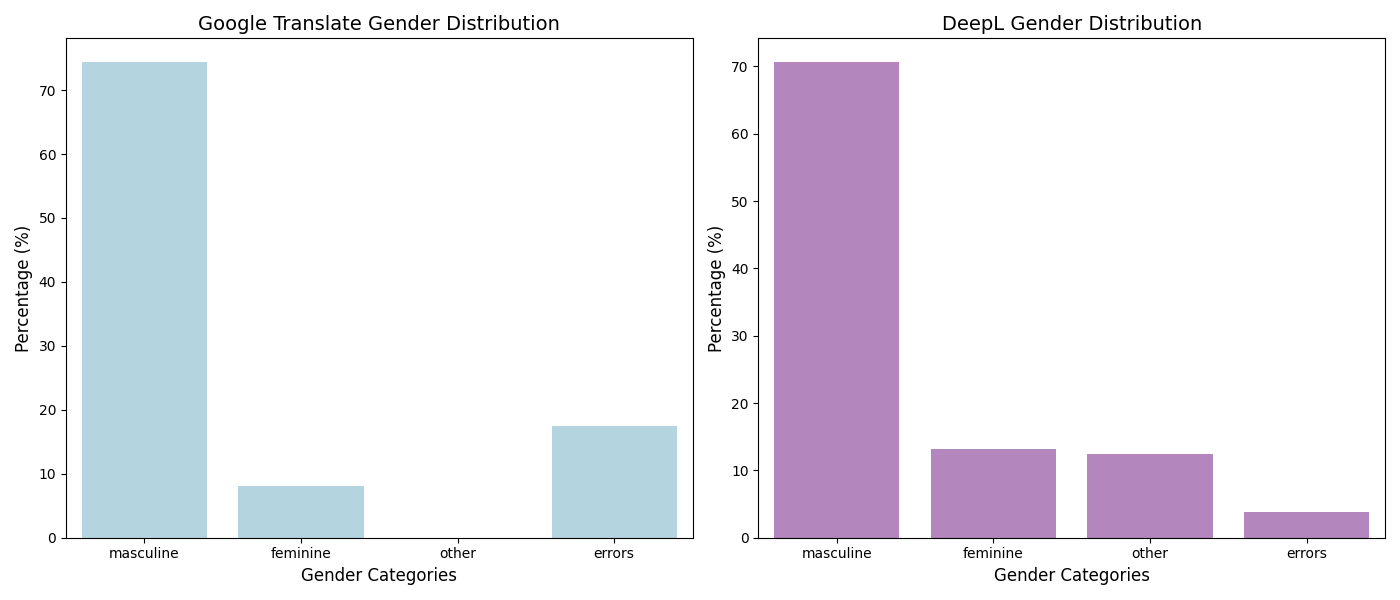}
  \caption{Gender distribution of translations for \textbf{gender-ambiguous} sentences by Google Translate and DeepL.}
  \label{fig:2}
\end{figure}

\begin{table}[htbp]
\centering
\begin{tabular}{@{}lcc@{}}
\toprule
\textbf{Gender Category} & \multicolumn{1}{l}{\textbf{Google Translate}} & \multicolumn{1}{l}{\textbf{DeepL}} \\ \midrule
masculine & 119 & 113 \\
feminine & 13 & 21 \\
other & - & 20 \\
errors & 28 & 6 \\ \midrule
\textbf{Total} & 160 & 160 \\ \bottomrule
\end{tabular}
\caption{Absolute counts for gender and error categories in translations by Google Translate and DeepL.}
\label{tab:1C-appendix}
\end{table}

\pagebreak
\section{Gender Bias in MT Systems: Occupational Stereotyping}
\label{sec:appendixE}

\begin{figure}[htbp]
  \centering
  \includegraphics[width=\columnwidth]{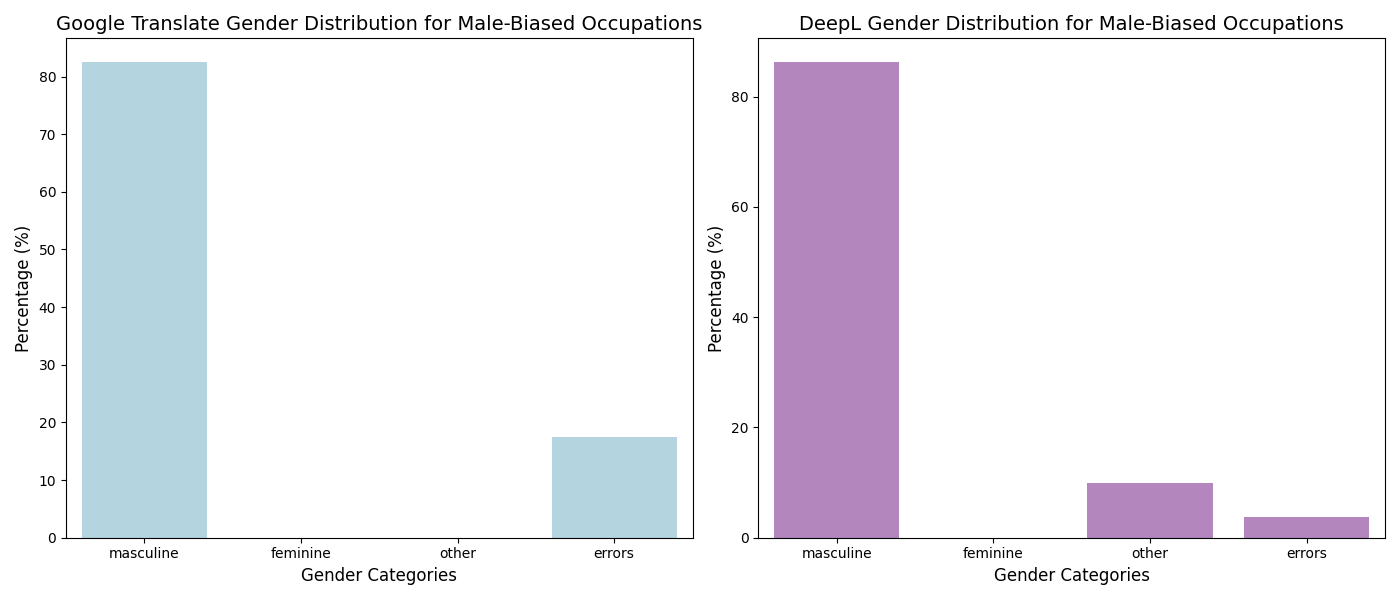}
  \caption{Gender distribution of translations for \textbf{stereotypically male occupations} in \textbf{gender-ambiguous} sentences, produced by Google Translate and DeepL.}
  \label{fig:3}
\end{figure}

\begin{table}[htbp]
\centering
\begin{tabular}{@{}lcc@{}}
\toprule
\textbf{Gender Category} & \multicolumn{1}{l}{\textbf{Google Translate}} & \multicolumn{1}{l}{\textbf{DeepL}} \\ \midrule
masculine & 66 & 69 \\
feminine & - & - \\
other & - & 8 \\
errors & 14 & 3 \\ \midrule
\textbf{Total} & 80 & 80 \\ \bottomrule
\end{tabular}
\caption{Absolute counts for gender distribution of translations for \textbf{stereotypically male occupations} in \textbf{gender-ambiguous sentences}, produced by Google Translate and DeepL.}
\label{tab:2C-appendix}
\end{table}

\begin{figure}[htbp]
  \centering
  \includegraphics[width=\columnwidth]{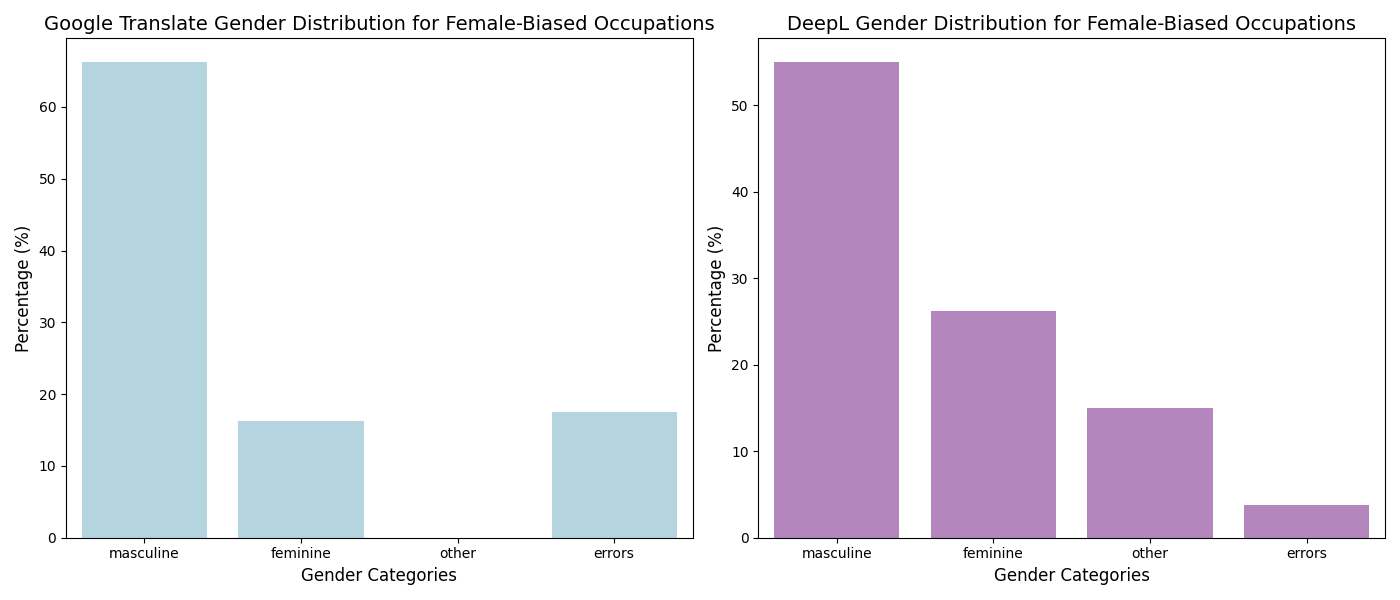}
  \caption{Gender distribution of translations for \textbf{stereotypically female occupations} in \textbf{gender-ambiguous} sentences, produced by Google Translate and DeepL.}
  \label{fig:4}
\end{figure}

\begin{table}[htbp]
\centering
\begin{tabular}{@{}lcc@{}}
\toprule
\textbf{Gender Category} & \multicolumn{1}{l}{\textbf{Google Translate}} & \multicolumn{1}{l}{\textbf{DeepL}} \\ \midrule
masculine & 53 & 44 \\
feminine & 13 & 21 \\
other & - & 12 \\
errors & 14 & 3 \\ \midrule
\textbf{Total} & 80 & 80 \\ \bottomrule
\end{tabular}
\caption{Absolute counts for gender distribution of translations for \textbf{stereotypically female occupations} in \textbf{gender-ambiguous sentences}, produced by Google Translate and DeepL.}
\label{tab:3C-appendix}
\end{table}

\pagebreak
\section{Gender Bias in MT Systems: Anti-Stereotypical Gender Assignments}
\label{sec:appendixF}

\begin{figure}[htbp]
  \centering
  \includegraphics[width=\columnwidth]{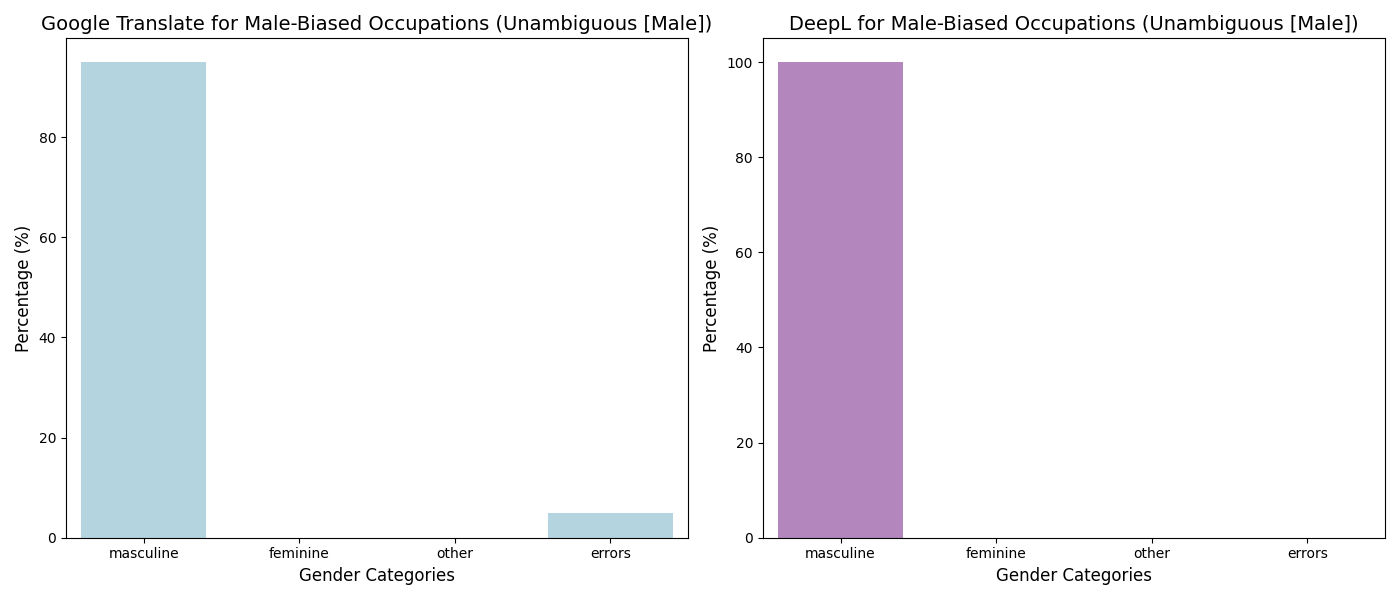}
  \caption{Gender distribution of translations for \textbf{stereotypically male occupations} in \textbf{masculine gender-unambiguous} sentences (stereotypical case), produced by Google Translate and DeepL.}
  \label{fig:5}
\end{figure}

\begin{table}[htbp]
\centering
\begin{tabular}{@{}lcc@{}}
\toprule
\textbf{Gender Category} & \multicolumn{1}{l}{\textbf{Google Translate}} & \multicolumn{1}{l}{\textbf{DeepL}} \\ \midrule
masculine & 19 & 20 \\
feminine & - & - \\
other & - & - \\
errors & 1 & - \\ \midrule
\textbf{Total} & 20 & 20 \\ \bottomrule
\end{tabular}
\caption{Absolute counts for gender distribution of translations for \textbf{stereotypically male occupations} in \textbf{masculine gender-unambiguous sentences} (stereotypical case), produced by Google Translate and DeepL.}
\label{tab:4C-appendix}
\end{table}

\pagebreak
\begin{figure}[htbp]
  \centering
  \includegraphics[width=\columnwidth]{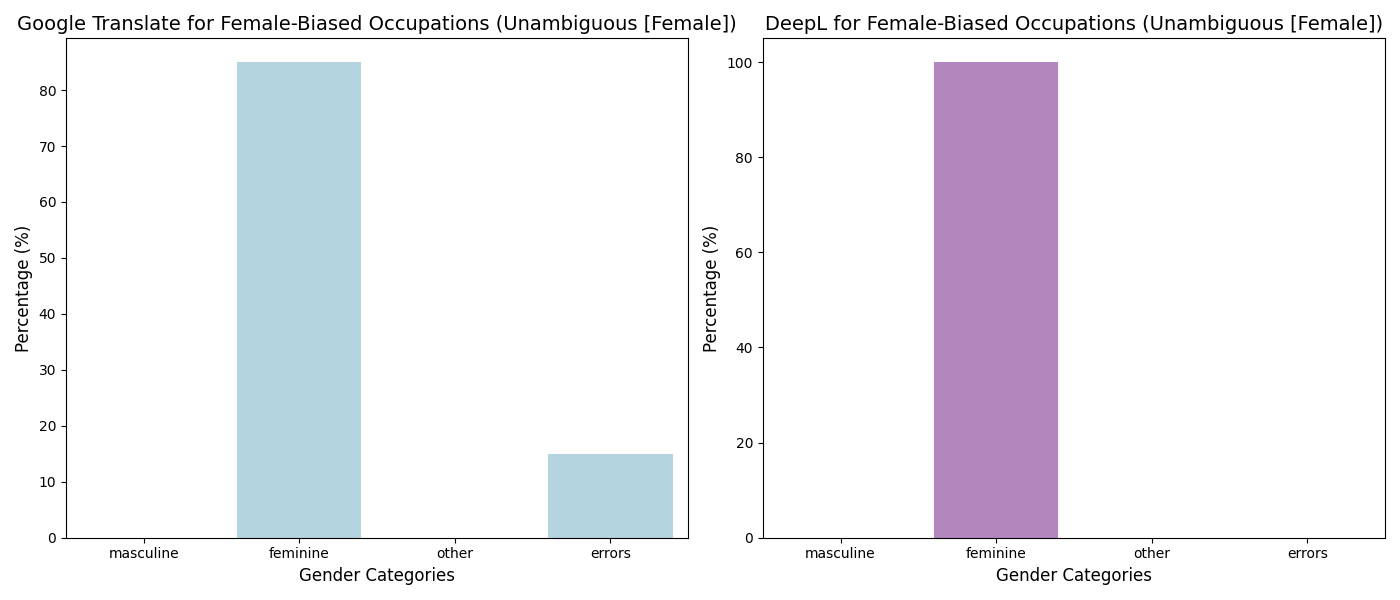}
  \caption{Gender distribution of translations for \textbf{stereotypically female occupations} in \textbf{feminine gender-unambiguous} sentences (stereotypical case), produced by Google Translate and DeepL.}
  \label{fig:6}
\end{figure}

\begin{table}[htbp]
\centering
\begin{tabular}{@{}lcc@{}}
\toprule
\textbf{Gender Category} & \multicolumn{1}{l}{\textbf{Google Translate}} & \multicolumn{1}{l}{\textbf{DeepL}} \\ \midrule
masculine & - & - \\
feminine & 17 & 20 \\
other & - & - \\
errors & 3 & - \\ \midrule
\textbf{Total} & 20 & 20 \\ \bottomrule
\end{tabular}
\caption{Absolute counts for gender distribution of translations of translations for \textbf{stereotypically female occupations} in \textbf{feminine gender-unambiguous sentences} (stereotypical case), produced by Google Translate and DeepL.}
\label{tab:5C-appendix}
\end{table}

\pagebreak
\begin{figure}[htbp]
  \centering
  \includegraphics[width=\columnwidth]{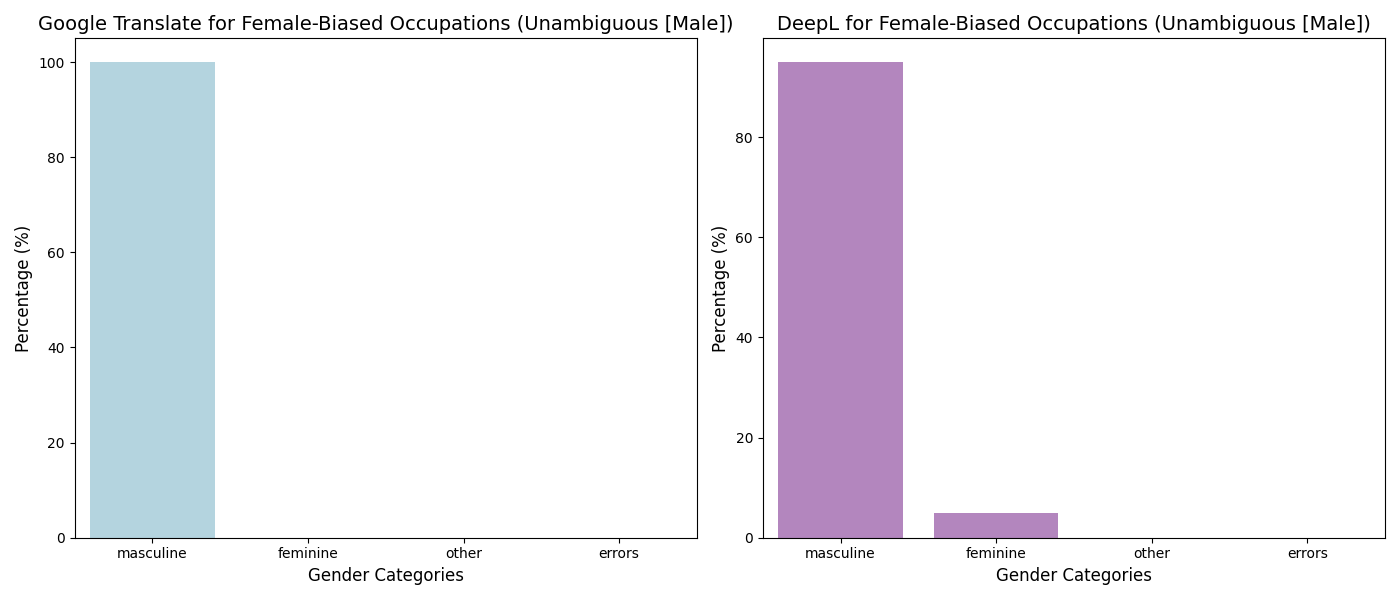}
  \caption{Gender distribution of translations for \textbf{stereotypically female occupations} in \textbf{masculine gender-unambiguous} sentences (anti-stereotypical case), produced by Google Translate and DeepL.}
  \label{fig:8}
\end{figure}

\begin{table}[htbp]
\centering
\begin{tabular}{@{}lcc@{}}
\toprule
\textbf{Gender Category} & \multicolumn{1}{l}{\textbf{Google Translate}} & \multicolumn{1}{l}{\textbf{DeepL}} \\ \midrule
masculine & 20 & 19 \\
feminine & - & 1 \\
other & - & - \\
errors & - & - \\ \midrule
\textbf{Total} & 20 & 20 \\ \bottomrule
\end{tabular}
\caption{Absolute counts for gender distribution of translations for \textbf{stereotypically female occupations} in \textbf{masculine gender-unambiguous sentences} (anti-stereotypical case), produced by Google Translate and DeepL.}
\label{tab:6C-appendix}
\end{table}

\pagebreak
\begin{figure}[htbp]
  \centering
  \includegraphics[width=\columnwidth]{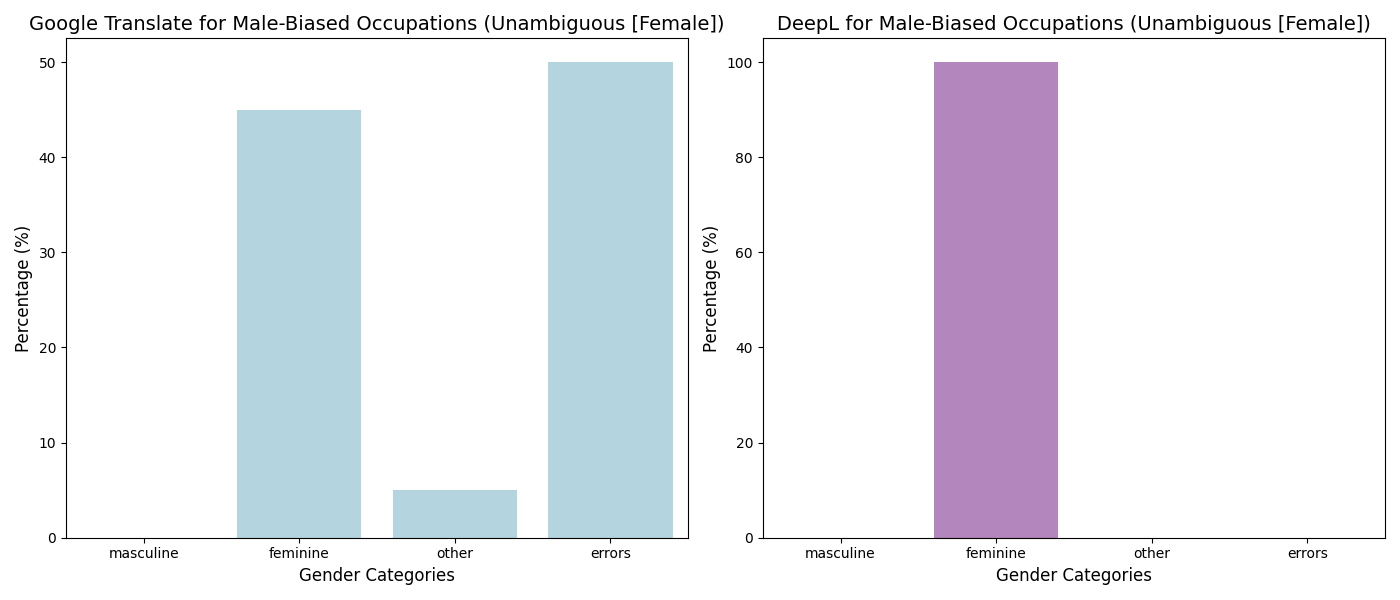}
  \caption{Gender distribution of translations for \textbf{stereotypically male occupations} in \textbf{feminine gender-unambiguous} sentences (anti-stereotypical case), produced by Google Translate and DeepL.}
  \label{fig:9}
\end{figure}

\begin{table}[htbp]
\centering
\begin{tabular}{@{}lcc@{}}
\toprule
\textbf{Gender Category} & \multicolumn{1}{l}{\textbf{Google Translate}} & \multicolumn{1}{l}{\textbf{DeepL}} \\ \midrule
masculine & - & - \\
feminine & 9 & 20 \\
other & 1 & - \\
errors & 10 & - \\ \midrule
\textbf{Total} & 20 & 20 \\ \bottomrule
\end{tabular}
\caption{Absolute counts for gender distribution of translations for \textbf{stereotypically male occupations} in \textbf{feminine gender-unambiguous sentences} (anti-stereotypical case), produced by Google Translate and DeepL.}
\label{tab:7C-appendix}
\end{table}

\pagebreak
\section{Results of Fischer's Exact Test}
\label{sec:appendixG}

\begin{table}[htbp]
  \centering
  \begin{tabular}{lll}
    \hline
    \textbf{Metric} & \textbf{Google Translate} & \textbf{DeepL} \\
    \hline
    Odds ratio & 0.0 & 0.0 \\
    $p$-value & 0.000143 & 2.105037e-07 \\
    \hline
  \end{tabular}
  \caption{\label{tab:1-appendix}
    Results of Fisher's exact test for Google Translate and DeepL, investigating whether there is a statistically significant association between the stereotype of the occupation (male- or female-biased) and the gender of the translation (e.g., masculine, feminine).
  }
\end{table}

\begin{table}[htbp]
  \centering
  \begin{tabular}{lll}
    \hline
    \textbf{Metric} & \textbf{Google Translate} & \textbf{DeepL} \\
    \hline
    Odds ratio & 0.0 & 0.0 \\
    $p$-value & 1.0 & inf \\
    \hline
  \end{tabular}
  \caption{\label{tab:2-appendix}
    Results of Fisher's exact test for Google Translate and DeepL, investigating whether there is a statistically significant difference between the anti-stereotypical and stereotypical groups when the gender is \textbf{masculine} and \textbf{unambiguous} in the English sentence.
  }
\end{table}

\begin{table}[htbp]
  \centering
  \begin{tabular}{lll}
    \hline
    \textbf{Metric} & \textbf{Google Translate} & \textbf{DeepL} \\
    \hline
    Odds ratio & 6.925925 & n/a \\
    $p$-value & 0.018701 & n/a \\
    \hline
  \end{tabular}
  \caption{\label{tab:3-appendix}
    Results of Fisher's exact test for Google Translate, investigating whether there is a statistically significant difference between the anti-stereotypical and stereotypical groups when the gender is \textbf{feminine} and \textbf{unambiguous} in the English sentence. For DeepL, both groups have exactly 20 feminine translations and 0 non-feminine translations, thus performing a Fisher's exact test is not meaningful.
  }
\end{table}

\pagebreak
\section{GPT-4o on Bias Mitigation}
\label{sec:appendixH}

\begin{figure}[htbp]
  \centering
  \includegraphics[width=\columnwidth]{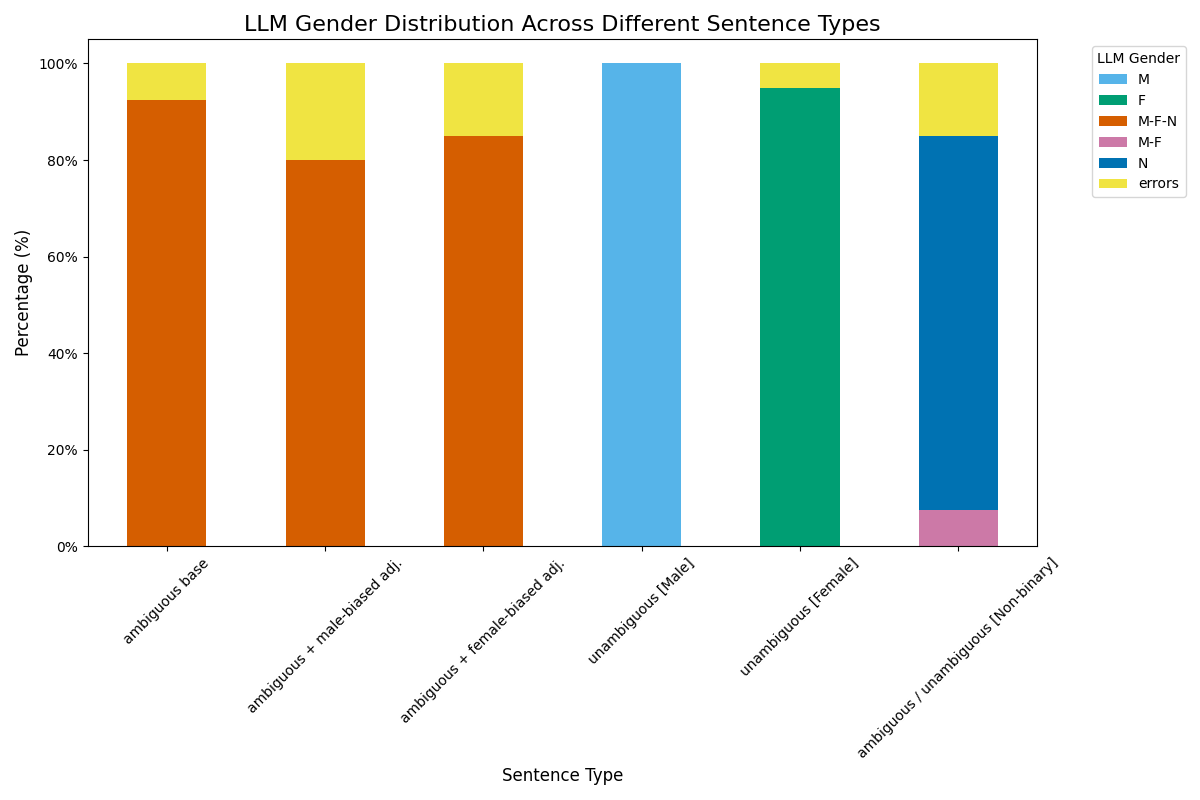}
  \caption{GPT-4o gender distribution across all sentence types.}
  \label{fig:12}
\end{figure}

\begin{figure}[htbp]
  \centering
  \includegraphics[width=\columnwidth]{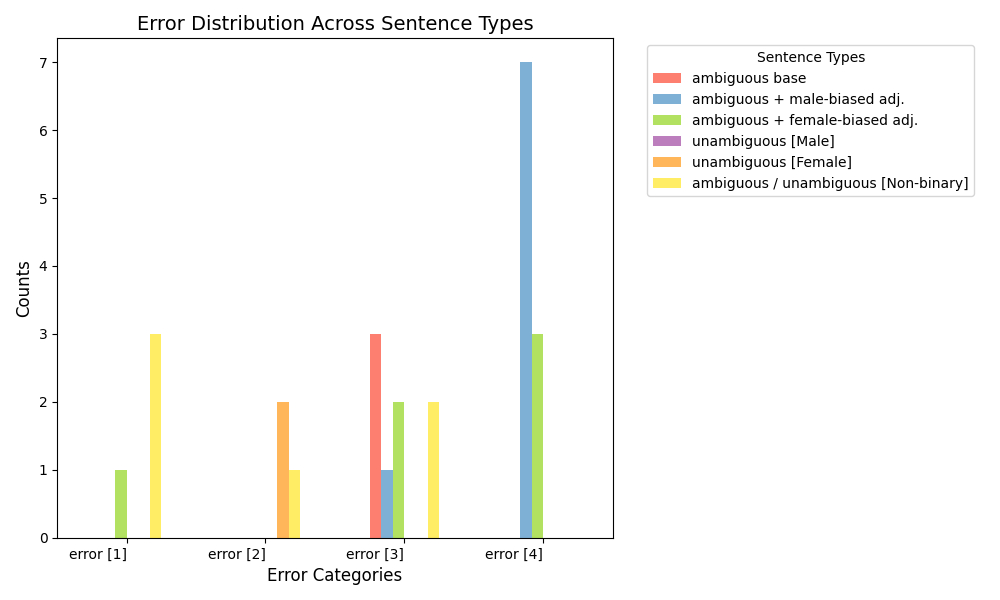}
  \caption{Error distribution of prompted GPT-4o across all sentence types.}
  \label{fig:15}
\end{figure}

\begin{table}[htbp]
\centering
\begin{tabular}{@{}|l|c|c|c|c|c|c|@{}}
\toprule
\multicolumn{1}{|c|}{\textbf{Type}} & \textbf{M} & \textbf{F} & \textbf{M-F-N} & \textbf{M-F} & \textbf{N} & \textbf{errors} \\ \midrule
ambiguous base & - & - & 37 (92.5\%) & - & - & 3 (7.5\%) \\ \midrule
\begin{tabular}[c]{@{}l@{}}ambiguous +\\ male-biased adj.\end{tabular} & - & - & 32 (80\%) & - & - & 8 (20\%) \\ \midrule
\begin{tabular}[c]{@{}l@{}}ambiguous + \\ female-biased adj.\end{tabular} & - & - & 34 (85\%) & - & - & 6 (15\%) \\ \midrule
\begin{tabular}[c]{@{}l@{}}unambiguous \\ {[}Male{]}\end{tabular} & 40 (100\%) & - & - & - & - & - \\ \midrule
\begin{tabular}[c]{@{}l@{}}unambiguous \\ {[}Female{]}\end{tabular} & - & 38 (95\%) & - & - & - & 2 (5\%) \\ \midrule
\begin{tabular}[c]{@{}l@{}}ambiguous / \\ unambiguous \\ {[}Non-binary{]}\end{tabular} & - & - & - & 3 (7.5\%) & 31 (77.5\%) & 6 (15\%) \\ \bottomrule
\end{tabular}
\caption{Absolute counts and percentages for gender categories and errors in translations by prompted GPT-4o.}
\label{tab:counts-gpt4o}
\end{table}

\begin{table}[htbp]
\centering
\begin{tabular}{@{}|l|l|l|@{}}
\toprule
\multicolumn{1}{|c|}{\textbf{\begin{tabular}[c]{@{}c@{}}Error\\ type\end{tabular}}} & \multicolumn{1}{c|}{\textbf{English source}} & \multicolumn{1}{c|}{\textbf{\begin{tabular}[c]{@{}c@{}}Translation by prompted\\ GPT-4o\end{tabular}}} \\ \midrule
3 & The mechanic finished the work. & \begin{tabular}[c]{@{}l@{}}\textgreek{Το άτομο που εργάζεται ως μηχανικός} \\ \textgreek{τελείωσε τη δουλειά.} (The person who \\ works as mechanic…)\end{tabular} \\ \midrule
3 & The sheriff finished the work. & \begin{tabular}[c]{@{}l@{}}\textgreek{Το άτομο που είναι σερίφης τελείωσε} \\ \textgreek{τη δουλειά.} (The person who is sheriff…)\end{tabular} \\ \midrule
3 & The CEO finished the work. & \begin{tabular}[c]{@{}l@{}}\textgreek{Το άτομο που είναι διευθύνων σύμβουλος} \\ \textgreek{τελείωσε τη δουλειά.} (The person who is \\ CEO…)\end{tabular} \\ \bottomrule
\end{tabular}
\caption{Error instances produced by prompted GPT-4o for the `ambiguous base' sentences.}
\label{tab:F1}
\end{table}

\begin{table}[htbp]
\centering
\begin{tabular}{@{}|l|l|l|@{}}
\toprule
\multicolumn{1}{|c|}{\textbf{\begin{tabular}[c]{@{}c@{}}Error\\ type\end{tabular}}} & \multicolumn{1}{c|}{\textbf{English source}} & \multicolumn{1}{c|}{\textbf{\begin{tabular}[c]{@{}c@{}}Translation by prompted\\ GPT-4o\end{tabular}}} \\ \midrule
3 & \begin{tabular}[c]{@{}l@{}}The suave mechanic finished \\ the work.\end{tabular} & \begin{tabular}[c]{@{}l@{}}\textgreek{Το κομψό άτομο που εργάζεται ως μηχανικός} \\ \textgreek{τελείωσε τη δουλειά.} (The suave person who \\ works as mechanic…)\end{tabular} \\ \midrule
4 & \begin{tabular}[c]{@{}l@{}}The grizzled mover finished \\ the work.\end{tabular} & \begin{tabular}[c]{@{}l@{}}\textgreek{Το άτομο που ασχολείται με τις μεταφορές} \\ \textgreek{τελείωσε τη δουλειά.}\end{tabular} \\ \midrule
4 & \begin{tabular}[c]{@{}l@{}}The arrogant sheriff finished \\ the work.\end{tabular} & \begin{tabular}[c]{@{}l@{}}\textgreek{Το άτομο της σερίφειας αρμοδιότητας} \\ \textgreek{τελείωσε τη δουλειά.}\end{tabular} \\ \midrule
4 & \begin{tabular}[c]{@{}l@{}}The grizzled developer finished \\ the work.\end{tabular} & \begin{tabular}[c]{@{}l@{}}\textgreek{Το άτομο που ασχολείται με τον} \\ \textgreek{προγραμματισμό τελείωσε τη δουλειά.}\end{tabular} \\ \midrule
4 & \begin{tabular}[c]{@{}l@{}}The jovial guard finished \\ the work.\end{tabular} & \begin{tabular}[c]{@{}l@{}}\textgreek{Το άτομο που φρουρεί τελείωσε τη} \\ \textgreek{δουλειά.}\end{tabular} \\ \midrule
4 & \begin{tabular}[c]{@{}l@{}}The shifty lawyer finished \\ the work.\end{tabular} & \begin{tabular}[c]{@{}l@{}}\textgreek{Το άτομο που ασκεί τη δικηγορία} \\ \textgreek{τελείωσε τη δουλειά.}\end{tabular} \\ \midrule
4 & \begin{tabular}[c]{@{}l@{}}The shifty writer finished \\ the work.\end{tabular} & \textgreek{Το άτομο που γράφει τελείωσε τη δουλειά.} \\ \midrule
4 & \begin{tabular}[c]{@{}l@{}}The rascally tailor finished \\ the work.\end{tabular} & \textgreek{Το άτομο που ράβει τελείωσε τη δουλειά.} \\ \bottomrule
\end{tabular}
\caption{Error instances produced by prompted GPT-4o for the `ambiguous + male-biased adj.' sentences.}
\label{tab:F2}
\end{table}

\begin{table}[htbp]
\centering
\begin{tabular}{@{}|l|l|l|@{}}
\toprule
\multicolumn{1}{|c|}{\textbf{\begin{tabular}[c]{@{}c@{}}Error\\ type\end{tabular}}} & \multicolumn{1}{c|}{\textbf{English source}} & \multicolumn{1}{c|}{\textbf{\begin{tabular}[c]{@{}c@{}}Translation by prompted\\ GPT-4o\end{tabular}}} \\ \midrule
1 & \begin{tabular}[c]{@{}l@{}}The bubbly carpenter finished \\ the work.\end{tabular} & \begin{tabular}[c]{@{}l@{}}\textgreek{Το ζωηρό άτομο που που ασχολείται} \\ \textgreek{με την ξυλουργική τελείωσε τη δουλειά.} \\ (repetition of ``\textgreek{που}''
 {[}who{]})\end{tabular} \\ \midrule
3 & \begin{tabular}[c]{@{}l@{}}The brunette mechanic finished\\ the work.\end{tabular} & \begin{tabular}[c]{@{}l@{}}\textgreek{Το άτομο με τα μελαχρινά μαλλιά που} \\ \textgreek{είναι μηχανικός τελείωσε τη δουλειά.}\end{tabular} \\ \midrule
3 & \begin{tabular}[c]{@{}l@{}}The vivacious CEO finished \\ the work.\end{tabular} & \begin{tabular}[c]{@{}l@{}}\textgreek{Το ζωηρό άτομο στη θέση του διευθύνοντος} \\ \textgreek{συμβούλου τελείωσε τη δουλειά.}\end{tabular} \\ \midrule
4 & \begin{tabular}[c]{@{}l@{}}The perky lawyer finished the \\ work.\end{tabular} & \begin{tabular}[c]{@{}l@{}}\textgreek{Το άτομο που ασκεί τη δικηγορία} \\ \textgreek{τελείωσε τη δουλειά.}\end{tabular} \\ \midrule
4 & \begin{tabular}[c]{@{}l@{}}The perky manager finished \\ the work.\end{tabular} & \begin{tabular}[c]{@{}l@{}}\textgreek{Το άτομο σε θέση διεύθυνσης τελείωσε} \\ \textgreek{τη δουλειά.}\end{tabular} \\ \midrule
4 & \begin{tabular}[c]{@{}l@{}}The alluring auditor finished\\ the work.\end{tabular} & \begin{tabular}[c]{@{}l@{}}\textgreek{Το άτομο που ελέγχει τελείωσε τη} \\ \textgreek{δουλειά.}\end{tabular} \\ \bottomrule
\end{tabular}
\caption{Error instances produced by prompted GPT-4o for the `ambiguous + female-biased adj.' sentences.}
\label{tab:F3}
\end{table}

\begin{table}[htbp]
\centering
\begin{tabular}{@{}|l|l|l|@{}}
\toprule
\multicolumn{1}{|c|}{\textbf{\begin{tabular}[c]{@{}c@{}}Error\\ type\end{tabular}}} & \multicolumn{1}{c|}{\textbf{English source}} & \multicolumn{1}{c|}{\textbf{\begin{tabular}[c]{@{}c@{}}Translation by prompted\\ GPT-4o\end{tabular}}} \\ \midrule
2 & The sheriff finished her work. & \begin{tabular}[c]{@{}l@{}}\textgreek{Ο σερίφης τελείωσε τη δουλειά της.} \\ (The male sheriff finished her work.)\end{tabular} \\ \midrule
2 & The farmer finished her work. & \begin{tabular}[c]{@{}l@{}}\textgreek{Ο αγρότης τελείωσε τη δουλειά της.} \\ (The male farmer finished her work.)\end{tabular} \\ \bottomrule
\end{tabular}
\caption{Error instances produced by prompted GPT-4o for the `unambiguous [Female]' sentences.}
\label{tab:F4}
\end{table}

\begin{table}[htbp]
\centering
\begin{tabular}{@{}|l|l|l|@{}}
\toprule
\multicolumn{1}{|c|}{\textbf{\begin{tabular}[c]{@{}c@{}}Error\\ type\end{tabular}}} & \multicolumn{1}{c|}{\textbf{English source}} & \multicolumn{1}{c|}{\textbf{\begin{tabular}[c]{@{}c@{}}Translation by prompted\\ GPT-4o\end{tabular}}} \\ \midrule
1 & \begin{tabular}[c]{@{}l@{}}The writer finished \\ their work.\end{tabular} & \begin{tabular}[c]{@{}l@{}}\textgreek{Η συγγραφέας/ο συγγραφέας τελείωσε} \\ \textgreek{τη δουλειά ατους.}\end{tabular} \\ \midrule
1 & \begin{tabular}[c]{@{}l@{}}The receptionist finished \\ their work.\end{tabular} & \begin{tabular}[c]{@{}l@{}}\textgreek{Η/Ο/Το ρεσεψιονίστ/ρεσεψιονίστρια/}\\ \textgreek{ρεσεψιονίστ τελείωσε τη δουλειά του/} \\ \textgreek{της/της.}\end{tabular} \\ \midrule
1 & \begin{tabular}[c]{@{}l@{}}The secretary finished \\ their work.\end{tabular} & \begin{tabular}[c]{@{}l@{}}\textgreek{Η/Ο/Το γραμματέας τελείωσε τη} \\ \textgreek{δουλειά του/της.}\end{tabular} \\ \midrule
2 & \begin{tabular}[c]{@{}l@{}}The editor finished their \\ work.\end{tabular} & \begin{tabular}[c]{@{}l@{}}\textgreek{Ο/Η συντάκτης/τρια τελείωσε τη} \\ \textgreek{δουλειά τους.}\end{tabular} \\ \midrule
3 & \begin{tabular}[c]{@{}l@{}}The mechanic finished \\ their work.\end{tabular} & \begin{tabular}[c]{@{}l@{}}\textgreek{Το άτομο που εργάζεται ως μηχανικός} \\ \textgreek{τελείωσε τη δουλειά.}\end{tabular} \\ \midrule
3 & \begin{tabular}[c]{@{}l@{}}The clerk finished their \\ work.\end{tabular} & \begin{tabular}[c]{@{}l@{}}\textgreek{Το άτομο που εργάζεται ως υπάλληλος} \\ \textgreek{τελείωσε τη δουλειά του.}\end{tabular} \\ \midrule
M-F & \begin{tabular}[c]{@{}l@{}}The developer finished \\ their work.\end{tabular} & \begin{tabular}[c]{@{}l@{}}\textgreek{Ο/Η προγραμματιστής/τρια τελείωσε} \\ \textgreek{τη δουλειά του/της.}\end{tabular} \\ \midrule
M-F & \begin{tabular}[c]{@{}l@{}}The farmer finished their \\ work.\end{tabular} & \begin{tabular}[c]{@{}l@{}}\textgreek{Το άτομο που δουλεύει στη φάρμα} \\ \textgreek{τελείωσε τη δουλειά του/της.}\end{tabular} \\ \midrule
M-F & \begin{tabular}[c]{@{}l@{}}The accountant finished \\ their work.\end{tabular} & \begin{tabular}[c]{@{}l@{}}\textgreek{Ο/Η λογιστής/λογίστρια τελείωσε} \\ \textgreek{τη δουλειά του/της.}\end{tabular} \\ \bottomrule
\end{tabular}
\caption{Error instances produced by prompted GPT-4o for the `ambiguous~/ unambiguous [Non-binary]' sentences.}
\label{tab:F5}
\end{table}

\pagebreak
\section{Gender Distribution of Google Translate, DeepL and GPT-4o in Unambiguous Cases}
\label{sec:appendixI}

\begin{figure}[htbp]
  \centering
  \includegraphics[width=\columnwidth]{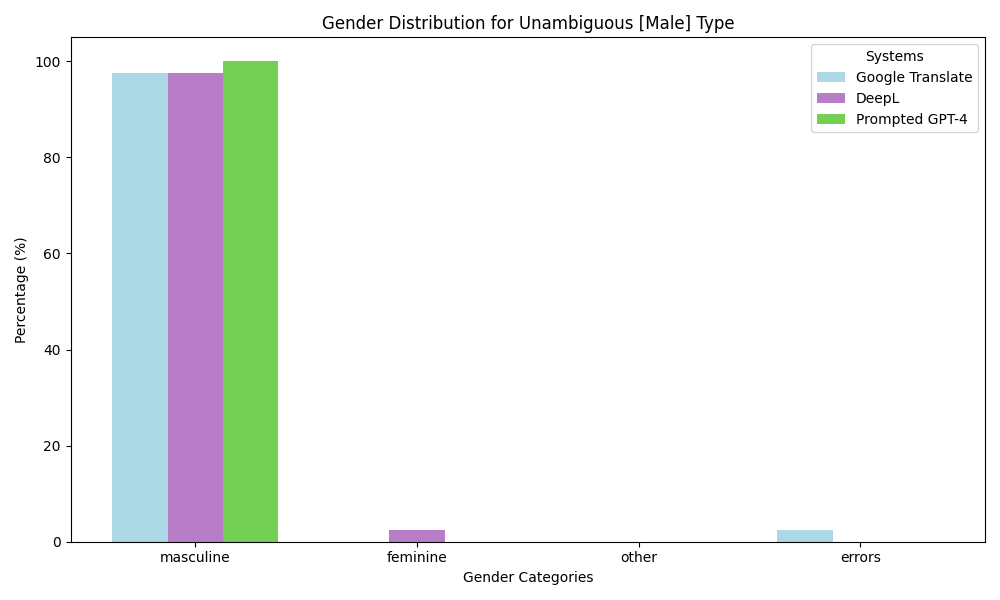}
  \caption{Gender distribution of translations for \textbf{masculine gender-unambiguous} sentences produced by Google Translate, DeepL, and (prompted) GPT-4o.}
  \label{fig:13}
\end{figure}

\begin{table}[htbp]
\centering
\begin{tabular}{@{}lccc@{}}
\toprule
\textbf{Gender Category} & \multicolumn{1}{l}{\textbf{Google Translate}} & \multicolumn{1}{l}{\textbf{DeepL}} & \multicolumn{1}{l}{\textbf{Prompted GPT-4o}} \\ \midrule
masculine & 39 & 39 & 40 \\
feminine & - & 1 & - \\
other & - & - & - \\
errors & 1 & - & - \\ \midrule
\textbf{Total} & 40 & 40 & 40 \\ \bottomrule
\end{tabular}
\caption{Absolute counts for gender distribution of translations for \textbf{masculine gender-unambiguous} sentences produced by Google Translate, DeepL and prompted GPT-4o.}
\label{tab:8C-appendix}
\end{table}

\pagebreak
\begin{figure}[htbp]
  \centering
  \includegraphics[width=\columnwidth]{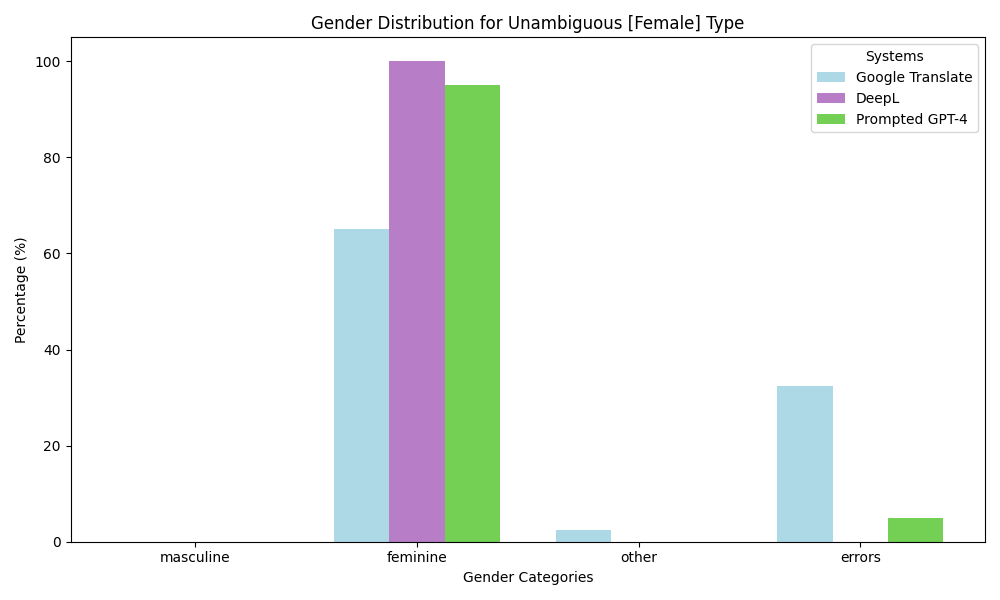}
  \caption{Gender distribution of translations for \textbf{feminine gender-unambiguous} sentences produced by Google Translate, DeepL, and (prompted) GPT-4o.}
  \label{fig:14}
\end{figure}

\begin{table}[htbp]
\centering
\begin{tabular}{@{}lccc@{}}
\toprule
\textbf{Gender Category} & \multicolumn{1}{l}{\textbf{Google Translate}} & \multicolumn{1}{l}{\textbf{DeepL}} & \multicolumn{1}{l}{\textbf{Prompted GPT-4o}} \\ \midrule
masculine & - & - & - \\
feminine & 26 & 40 & 38 \\
other & 1 & - & - \\
errors & 13 & - & 2 \\ \midrule
\textbf{Total} & 40 & 40 & 40 \\ \bottomrule
\end{tabular}
\caption{Absolute counts for gender distribution of translations for \textbf{feminine gender-unambiguous} sentences produced by Google Translate, DeepL and prompted GPT-4o.}
\label{tab:8C-appendix}
\end{table}

\end{document}